%% file: main.tex
\documentclass[10pt,twocolumn,letterpaper]{article}

\usepackage[pagenumbers]{cvpr} %

\input{preamble}

\definecolor{cvprblue}{rgb}{0.21,0.49,0.74}
\usepackage[pagebackref,breaklinks,colorlinks,allcolors=cvprblue]{hyperref}

\title{VCURF: Virtual Camera-based Uncertainty of Radiance Fields}

\author{Liyan Chen \quad Nathaniel Burgdorfer \quad Philippos Mordohai\\
Stevens Institute of Technology\\
{\tt\small \{lchen39, nburgdor, pmordoha@stevens.edu\}}
}
\begin{document}
\maketitle

\begin{abstract}
\input{sec/abstract}

\end{abstract}

\section{Introduction}
\label{sec:intro}
\input{sec/introduction}

\section{Related Work}
\label{sec:related_works}
\input{sec/related}

\section{Method}
\label{sec:method}

\input{sec/method}

\section{Experimental Results}
\label{sec:experiments}
\input{sec/experiments}

\section{Discussion and Future work}
\label{sec:discussion}
\input{sec/discussion}

\section{Conclusion}
\label{sec:conclusion}

\input{sec/conclusion}

{
    \small
    \bibliographystyle{ieeenat_fullname}
     \bibliography{bib/ActiveMapping,bib/nerf,bib/other}
}

\input{sec/X_suppl}

\end{document}

%% file: sec/abstract.tex
Radiance fields, implemented with either implicit (NeRF) or explicit (Gaussian Splatting) representations, are advancing the state of the art in novel view synthesis at a rapid pace. Even though the rendered views they generate are often compelling, they are not free of errors. In this paper, we propose a new approach for pixel-wise uncertainty quantification based on measuring the inconsistencies among renderings by the radiance field model in virtual cameras sampled near the target viewpoint. We named our approach VCURF for Virtual Camera-based Uncertainty of Radiance Fields. VCURF treats the radiance field model as a black box, only assuming that it is capable of rendering color and depth on demand. This property makes our approach applicable to both NeRF and GS models without any modification. Our experiments on a combination of datasets, radiance field models and baselines demonstrate VCURF's effectiveness in pixel-wise uncertainty estimation. We conclude the paper with findings that question the way view selection is tackled by the majority of the current literature.

%% file: sec/introduction.tex
Neural Radiance Fields (NeRFs) \cite{mildenhall2020nerf} have had tremendous impact on Novel View Synthesis (NVS). This is due to their ability  to synthesize plausible images without striking artifacts under adverse imaging conditions and  errors in camera geometry estimation. More recently, 3D Gaussian Splatting (3DGS) \cite{kerbl20233dgs} emerged as an alternative to NeRF. Unlike NeRF, which relies on implicit representations, Gaussian Splatting employs an explicit representation in the form of a set of 3D Gaussian splats. Following recent literature, we will use the term \textit{Radiance Fields} (RF) to describe both. (We will also use GS to denote Gaussian Splatting in general, and 3DGS for the method of Kerbl et al. \cite{kerbl20233dgs}.)

Despite their success, RF-based maps are not perfect in practice: limited views, appearance challenges, dynamics, and pose errors can cause local failures in geometry and appearance. This motivates pixel-wise uncertainty estimation that predicts the \emph{expected} rendering error without ground truth.
Such estimates provide useful signals for identifying unreliable regions, filtering artifacts, and assessing local reconstruction quality. They have also been widely used as proxies for information gain in next-best-view (NBV) selection and active reconstruction.

However, these two uses should be distinguished. Pixel-wise uncertainty characterizes where the current model is likely to be wrong, whereas the utility of a candidate view additionally depends on whether it contributes novel and non-redundant observations relative to the existing training set. A highly uncertain view is therefore not necessarily the view that will most improve the RF after retraining. We revisit this distinction experimentally after establishing the accuracy of our uncertainty estimator.

In practice, we seek an uncertainty module that is easy to deploy as a \emph{plug-in}, is \emph{training-free}, \emph{memory-friendly}, and  avoids modifying or accessing internal RF parameters, which simplifies integration across diverse backbones. 
In this paper, we use \textit{black-box} to denote 
estimators that require only outputs and no knowledge of model parameters, and \textit{white-box} to denote methods that require modifying the RF model and/or introducing additional training objectives.

\begin{figure}[t]
    \centering
    {\includegraphics[width=0.95\columnwidth]{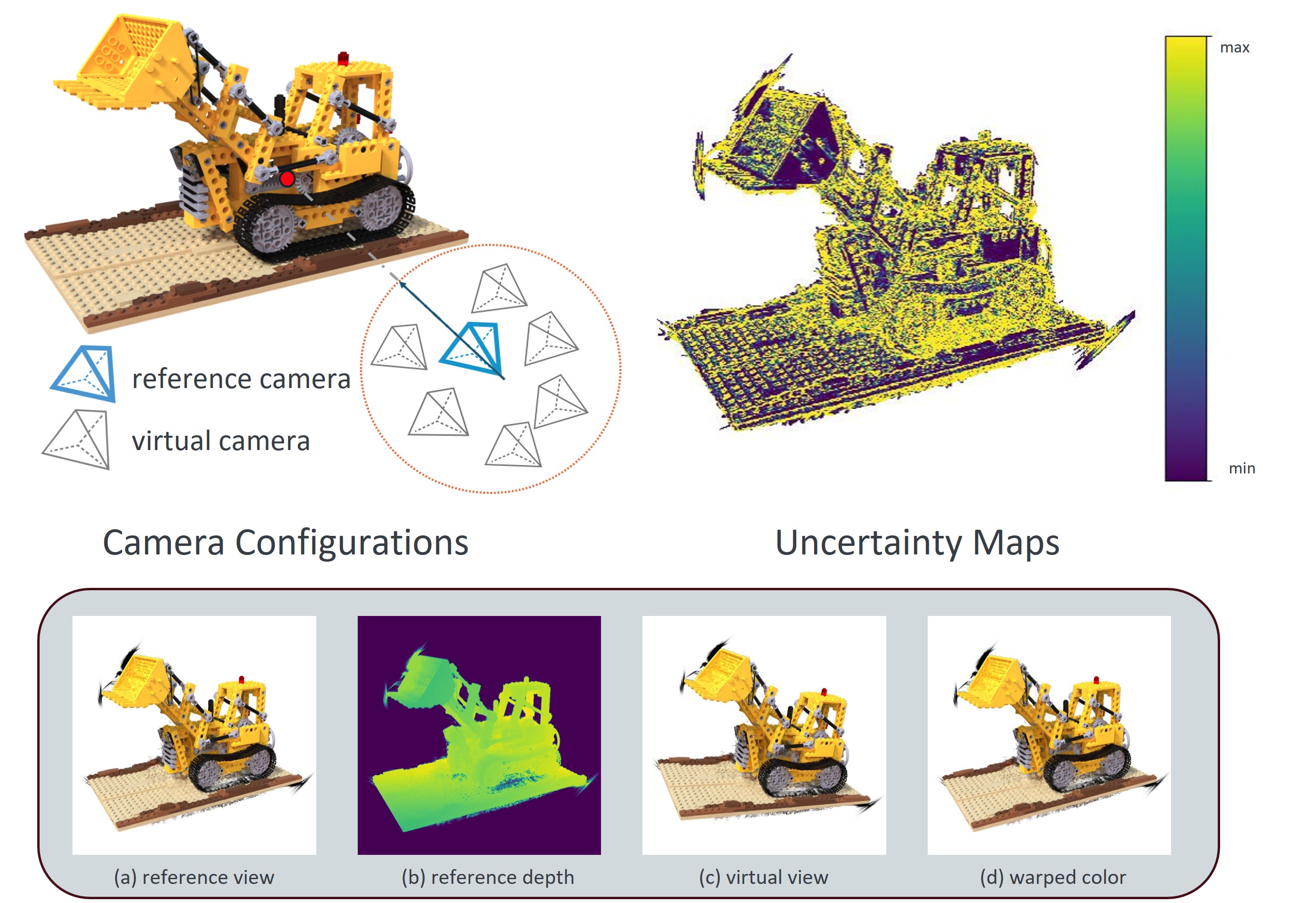}
    }
    \caption{Overview of our approach. Top Left: virtual cameras are sampled around the reference camera. Top Right: uncertainty predicted by \methodname. Bottom: (a) real image, (b-c) depth and color rendered by the RF model, and (d) rendered color image warped from a virtual camera to the reference.}
    \label{fig:overview}
    \vspace{-15pt}
\end{figure}

Motivated by these constraints, we propose \textit{Virtual-Camera-based Uncertainty of Radiance Fields (\methodname)}, a \emph{black-box} pixel-wise uncertainty estimator applicable to both implicit and explicit RFs. \methodname is based on the principle that \emph{inconsistency 
indicates uncertainty}: it perturbs the \emph{viewpoint} via a set of \textit{virtual cameras} around the target view, warps renderings across views, and derives uncertainty from discrepancies in warped color and depth (Fig.~\ref{fig:overview}). 
Compared to uncertainty estimators that require stochastic inference, posterior approximations, or additional model instances/components (e.g., ensembles, dropout, entropy, Bayesian/variational methods)~\cite{sunderhauf2023density,sun2024ph,zhan2022activermap,xue2024neural,lyu2024manifold}, \methodname is a lightweight plug-in estimator that queries only renderings to virtual cameras, avoiding retraining and the storage overhead of maintaining multiple model instances while still producing accurate pixel-wise uncertainty.

Two recent methods rely on Fisher Information to estimate uncertainty: Bayes' Rays \cite{goli2024bayes} for NeRF and FisherRF \cite{jiang2024fisherrf} for GS. Both approaches derive uncertainty from a Hessian matrix and impose the Laplace approximation to make computation tractable. These formulations are elegant, but the range within which the approximations are valid may be limited. \methodname, on the other hand, does not impose any restrictions on the posterior of the model. As a result, uncertainty cannot be computed in closed form, given model parameters, but we are able to measure the effects of relatively large perturbations in camera pose. 
Our experimental results show higher accuracy in uncertainty prediction compared to FisherRF.

The main contributions of our paper are:
\begin{itemize}
\item a new approach for estimating pixel uncertainty of implicit and explicit RF models based on virtual-camera cross-view consistency;
\item the \methodname algorithm, a \emph{training-free, memory-friendly} plug-in uncertainty estimator that treats the RF model as a black box (no model modification and no access to parameters) and does not require storing multiple models;
\item extensive experiments on multiple datasets comparing \methodname with various uncertainty estimators on an implicit and an explicit RF model.
\end{itemize}

%% file: sec/related.tex
As the literature on Radiance Fields (RFs) expands rapidly, providing a
comprehensive survey is beyond the scope of this paper. We refer readers to
recent overviews \cite{bao2025_3dgs_survey,dalal_2024_gaussian,fei2024survey}
and focus on uncertainty estimation for NeRF and Gaussian Splatting (GS).
Klasson et al. \cite{klasson2024sources} evaluate uncertainty predictors for
both representations and categorize common sources of uncertainty as noise,
occlusions, outliers, and imprecise camera poses. Their study evaluates uncertainty
quantification due to each source separately without considering identifying the sources.

\paragraph{Uncertainty estimation for NeRF.}
A common approach augments the reconstruction objective with a negative
log-likelihood (NLL) term \cite{kendall2017uncertainties}. NeRF-W
\cite{martin2021nerfwild} uses this formulation to account for transient
occluders and other outliers, while Neu-NBV \cite{jin2023neu}, ActiveNeRF
\cite{pan2022activenerf}, the Colorized Surface Voxel approach
\cite{kim2024active}, and NARUTO \cite{feng2024naruto} use related predictive
uncertainty formulations in active reconstruction. Neural Visibility Fields
\cite{xue2024neural} jointly model visibility and uncertainty in a Bayesian
formulation that also incorporates an NLL objective. Entropy along a ray has
likewise been used as an uncertainty signal
\cite{zhan2022activermap,lee2022uncertainty,yan2023activeimplicit}, while NeurAR
\cite{ran2023neurar} uses transmittance-weighted rendering variance but
requires ground-truth depth during training.

Ensembles provide another widely used estimate of epistemic uncertainty.
S\"underhauf et al. \cite{sunderhauf2023density} measure variation among RGB
renderings and incorporate predicted density, outperforming naive ensembles.
He et al. \cite{he_chaudhari2024active} form distributions over color, depth,
voxel occupancy, and semantic labels. Dropsembles
\cite{susmelj2025uncertainty} construct an ensemble by fine-tuning several
dropout-thinned instances on noisy inputs, while Yan et al.
\cite{yan2023activeneuralmapping} perturb network weights during exploration
and use the resulting rendering variation to identify uncertain regions.
Monte Carlo Dropout \cite{gal2016dropout} can also be viewed as an efficient
mechanism for constructing an implicit ensemble and is widely used as an
uncertainty baseline
\cite{shen2021stochastic,shen2022conditional,sunderhauf2023density,yan2023activeneuralmapping}.

Rather than maintaining a finite set of models, Conditional-Flow NeRF
(CF-NeRF) \cite{shen2022conditional} learns a distribution over possible RFs
using conditional normalizing flows and estimates uncertainty from multiple
samples. Shen et al. \cite{shen2024estimating} instead combine an NLL-based
prediction with a 3D uncertainty field that distinguishes observed from
unobserved regions.  Zeng et al. \cite{zeng2023efficient} learn an approximation to a volumetric information-gain field. Bayes' Rays \cite{goli2024bayes} defines a spatial
perturbation field and estimates uncertainty through the Hessian of the
posterior over network parameters: regions are considered uncertain when large
perturbations produce little change in the objective. Evidential Neural Radiance Fields \cite{duan2026evidential} is a probabilistic approach for distinguishing between epistemic and aleatoric uncertainty.

\paragraph{Uncertainty estimation for Gaussian Splatting.}
WildGaussians \cite{kulhanek2024wildgaussians} uses an NLL loss to predict
uncertainty caused by occlusions, dynamic objects, and illumination changes.
ActiveGS \cite{jin2025activegs} derives the uncertainty of each splat from the
density and angular diversity of its observing cameras, while Gauss-MI
\cite{xie2025GauSS-MI} associates reliability variables with the splats and
uses them to estimate information gain. Variational formulations have been
introduced by Savant et al. \cite{savant2025modeling}, who incorporate an
AUSE-based term into the training objective, and by Li et al.
\cite{li2024variational}, who use multi-scale analysis. Manifold sampling
\cite{lyu2024manifold} provides a differentiable uncertainty formulation based
on variational inference. RT-GuIDE \cite{tao2024rt} uses the magnitude of
Gaussian-parameter updates as its uncertainty signal, whereas Gottwald et al.
\cite{gottwald2025uncertainty} attach visibility and training-error features to
the primitives and train a regressor to predict pixel-wise uncertainty. Xue et al. \cite{xue2026uncertainty} estimate an anisotropic volumetric visibility field for explicit RFs, from which uncertainty can be derived.

FisherRF \cite{jiang2024fisherrf} uses Fisher information to estimate the
information carried by a view or pixel about the underlying RF. It obtains a
tractable estimate from a diagonal Laplace approximation to the Hessian of the
rendering objective. This approximation is particularly suitable for
representations with structured local parameters, such as 3DGS
\cite{kerbl20233dgs} and Plenoxels \cite{fridovich2022plenoxels}. PUP 3D-GS
\cite{hanson2024pup} also computes Fisher information for Gaussian primitives,
but uses a block-wise rather than diagonal approximation and evaluates
uncertainty indirectly through pruning.

\paragraph{Representation-agnostic estimators.}
Only a few uncertainty estimators can directly work with both implicit and explicit
RFs. FisherRF is applicable to both, although its approximation can be poorer
for fully connected implicit models. Ewen et al. \cite{ewen2025these} use low-
and high-order moments of rendered color, depth, and semantic labels.
PH-Dropout \cite{sun2024ph} inserts dropout into either a trained NeRF or a GS
model, increases the dropout ratio while preserving training-view accuracy,
and measures uncertainty from the variance of repeated renderings. Like our
method, WarpRF \cite{safadoust2026warprf} estimates uncertainty through
cross-view consistency without modifying the RF architecture. It renders from
the training poses, warps those renderings to a reference view using
predicted depth, and measures their discrepancy. VCURF instead samples virtual
cameras locally around every reference viewpoint, yielding controlled and
comparable camera configurations without requiring nearby training views.

\paragraph{Uncertainty and Active View Selection.}
Uncertainty is widely used as a proxy for information gain in active RF
reconstruction. Neu-NBV \cite{jin2023neu}, ActiveNeRF
\cite{pan2022activenerf}, NARUTO \cite{feng2024naruto}, Neural Visibility
Fields \cite{xue2024neural,xue2026uncertainty}, FisherRF \cite{jiang2024fisherrf}, and manifold
sampling \cite{lyu2024manifold}, among others, rank candidate observations
using predicted uncertainty or an approximation to the expected reduction in
model uncertainty. Related methods use ray entropy
\cite{zhan2022activermap,lee2022uncertainty,yan2023activeimplicit}, learned
volumetric information gain \cite{zeng2023efficient}, or primitive reliability
\cite{xie2025GauSS-MI}. Although their formulations differ, these criteria
primarily characterize where the current model is uncertain, unreliable, or
sensitive to additional observations.

A complementary family of methods selects views using global information about
the current observations. Geometric criteria \cite{kopanas2023improving} and
camera-distribution or redundancy heuristics
\cite{xiao2024nerfdirector,wang2024redundancy} explicitly favor views that are
poorly represented by the existing training set. PRIMU
\cite{gottwald2026primu}, for example, evaluates active view selection under
the FisherRF protocol and reports that a criterion based directly on
training-view coverage outperforms baselines that estimate uncertainty or
approximate information gain.

These two families capture related but distinct properties. Pixel-wise
uncertainty approximates local predicted error under the current RF,
whereas the utility of acquiring a candidate view also depends on whether it
provides novel and non-redundant observations relative to the current training
set. Consequently, a view with high predicted error need not produce
the largest improvement after it is added to the training data. Conversely,
coverage-only criteria can overlook image-dependent challenges such as
occlusions, concavities, and spatially varying levels of detail. VCURF focuses
on accurate pixel-wise rendering uncertainty. Our active-view-selection
experiment and subsequent discussion revisit whether training error alone is
equivalent to view utility.

%% file: sec/method.tex
In this section, we introduce our approach, Virtual Camera-based Uncertainty of Radiance Fields (\methodname), for estimating the uncertainty of rendering by an RF using virtual cameras, and
present how it can be implemented 
over both explicit and implicit rendering pipelines. 
We start with the geometry of \methodname,
and then introduce the core of this paper, which is a black-box technique for quantifying uncertainty by warping depth and color information from the virtual cameras to the reference camera.

\subsection{Virtual Camera Configuration}\label{sec:virtual}

Given a trained NeRF or optimized GS, concisely an RF denoted by $\mathbf{\phi}$, we wish to estimate the uncertainty of a novel view that would be rendered by it, without access to an image from that viewpoint. We begin by constructing a set of virtual cameras surrounding the camera at the viewpoint of interest, which we refer to as the \textit{reference camera}. Throughout, we use $N$ for the number of virtual cameras sampled within a sphere centered at the reference camera as shown in Fig.~\ref{fig:overview}.

Given a reference camera with projection matrix $K_{r}[R_{r} \: t_{r}]$, where $K_{r}$ is the intrinsic calibration matrix and $[R_{r} \: t_{r}]$ is the $3\times 4$ matrix of extrinsics, we query the trained RF, $\mathbf{\phi}$, to render the \textit{reference depth map}  $D_{r}$. (In general, we use capital letters for sets or maps, such as the depth map $D$ and lower-case symbols for their members, such as the depth $d_i$ at pixel $i$.) We use $D_r$ to parameterize virtual camera placement.

Virtual cameras should be placed in a way that balances (i) overlap with the reference so that points are visible in multiple views and (ii) sensitivity to depth which requires larger baselines. The first objective is achieved by ensuring that virtual cameras are 
oriented similarly to the reference camera and surround it,
so that even boundary pixels of the reference are still likely to project inside some virtual images. The second objective is achieved by sampling virtual cameras in a sufficiently large space. Randomly sampling a sufficient number of cameras achieves this balance better than manual specification; see supporting experiments in the supplement.

The intrinsics of the virtual cameras are set equal to those of the reference, $K_v = K_r$. 
To set the extrinsics, we randomly sample a center for each virtual camera in a sphere centered around the reference view. We set the radius of the sphere $r_s$ as a fraction $p_1$ of the median depth $d_{med}$ in $D_r$. We also require the distances between camera centers to be greater or equal to $p_{2}r_s$, where $p_2$ is a constant less than 1, to prevent the generation of redundant cameras. Denoting camera centers by $Q$, we sample a set of $N$ virtual cameras:
\begin{align}
&p_{2}r_s \leq \|Q_{v_i} - Q_r\|_2 \leq r_s  \\
    &\|Q_{v_i} - Q_{v_j}\|_2 \geq p_{2}r_s \quad \text{for all $i$ and $j$}
\end{align}
We then orient each virtual camera so that it points to the centroid $O$ of the 3D points backprojected from $D_r$. To complete the camera coordinate system, we use the average of the up vectors of all \textit{training cameras} as the world-up reference, $\overrightarrow{u}$, which combined with the viewing direction can yield an orthonormal rotation matrix, $R_{v}$. 

\begin{figure}[tb]
    \centering
    \includegraphics[width=0.7\columnwidth]{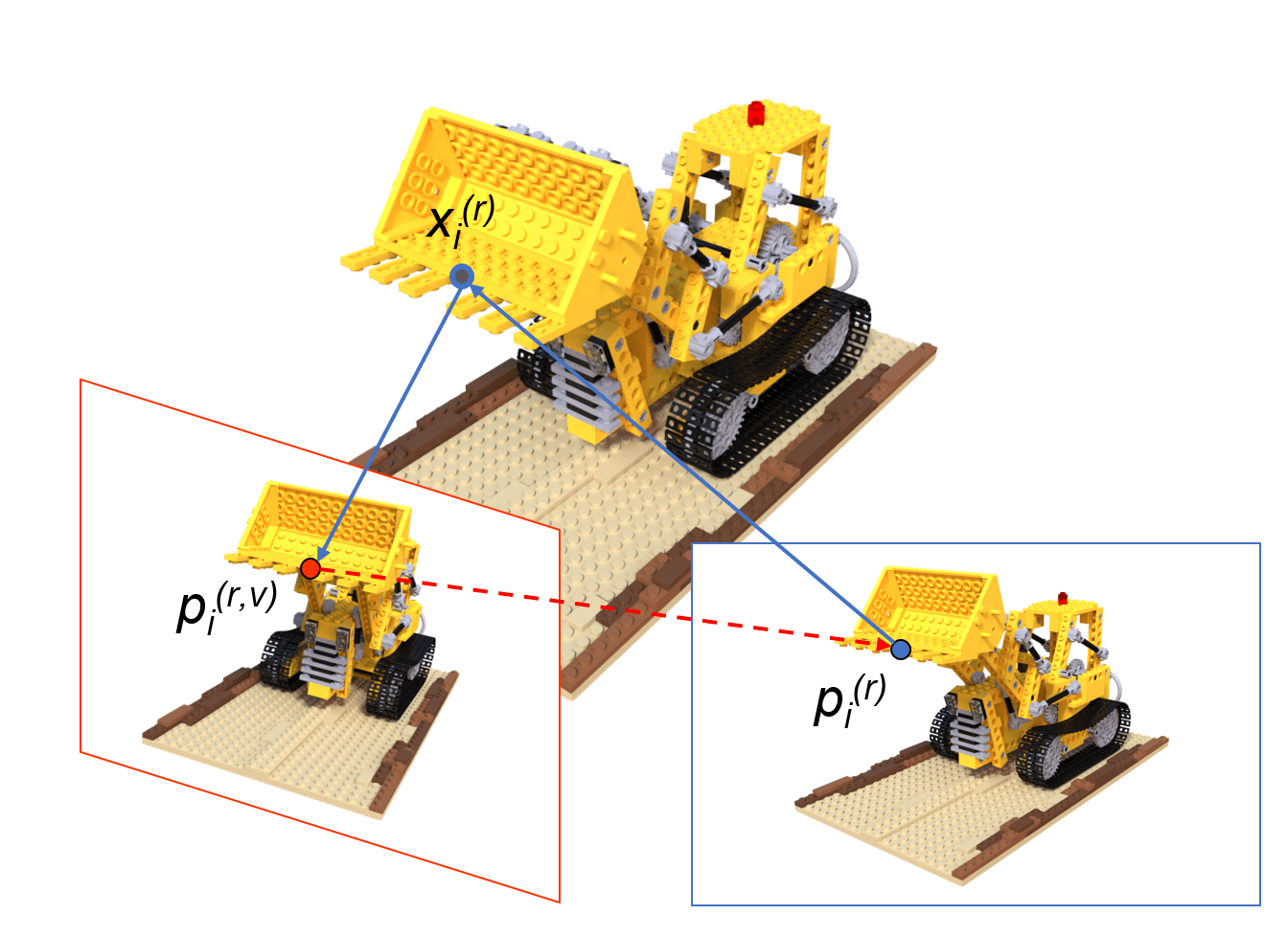}
    \caption{Pixel $p_i^{(r)}$ is projected  to virtual view $v$ via the reconstructed 3D point $x_i^{(r)}$.
    Color and depth are obtained by bilinear interpolation at the projected point $p_i^{(r,v)}$,
    and then associated back to the reference pixel $p_i^{(r)}$.}
    \label{fig:backwarp}
    \vspace{-10px}
\end{figure}

\subsection{Warping Depth and Color}\label{sec:depthwarping}

We render an image $C_v$ and a depth map $D_v$ to a virtual camera $v$ using the RF model $\mathbf{\phi}$. We then measure consistency with the reference view, for which we have obtained $C_r$ and $D_r$. 
For a pixel $p^{(r)}_i$ with coordinates $(u^{(r)}_i, v^{(r)}_i)$ and depth $d^{(r)}_i$ in the reference depth map, we can obtain the corresponding 3D point in the reference camera coordinate system $x^{(r)}_i$ via backprojection:

\begin{equation}
    x^{(r)}_i = K_{r}^{-1}{\left[u^{(r)}_i\: v^{(r)}_i \: 1 \right]}^{T} d^{(r)}_i \label{eq:backproj}
\end{equation}

\noindent The 3D point is then transformed to the world coordinate system and projected to virtual camera $v$. For efficiency, these transformations can be combined to obtain the relative transformation from the reference camera $r$ to camera $v$. Using the $4\times 4$ matrices of world to camera rigid transformations: $T_{r,v} = T_{v}T_r^{-1}$.
We also obtain the projection matrix $[R_{r,v} \: t_{r,v}]$ that projects 3D points from the coordinate system of camera $r$ to camera $v$ from the first three rows of $T_{r,v}$.
The 3D point $x^{(r)}_i$ will be projected to a pixel in camera $v$:
\begin{equation}
    p^{(r,v)}_i \sim K_{v}\left[R_{r,v} \: t_{r,v}\right] x^{(r)}_i \label{eq:proj}
\end{equation}
\noindent where the superscript $^{(r,v)}$ denotes a variable that has been mapped from the reference view $r$ to virtual view $v$.
We obtain the depth $d^{(r,v)}_i$ and color $c^{(r,v)}_i$ of $p^{(r,v)}_i$ by bilinear interpolation. 
We can then associate the color $c^{(r,v)}_i$ with the original pixel $p_i^{(r)}$ in the reference view. To do the same for the depth, we have to backproject from the depth map of the virtual camera to 3D and measure the depth of the backprojected 3D point with respect to the reference camera. We denote this by $d_i^{(v,r)}$ since it is the depth of 3D point $x_i^{r}$ estimated for virtual camera $v$ and warped to the reference camera $r$. See Fig.~\ref{fig:backwarp}.

\subsection{Black-Box Uncertainty Estimation}\label{sec:blackbox}

We can now introduce our black-box uncertainty quantification approach. Unlike other black-box methods, such as ensembles, 
our approach does not require additional training or optimization time, nor does it require additional storage for multiple instances of the model.
The additional cost of \methodname is rendering operations for a few virtual views, making it a more efficient solution than ensembles.

\begin{figure}[t]
    \centering
    \includegraphics[width=0.55\columnwidth]{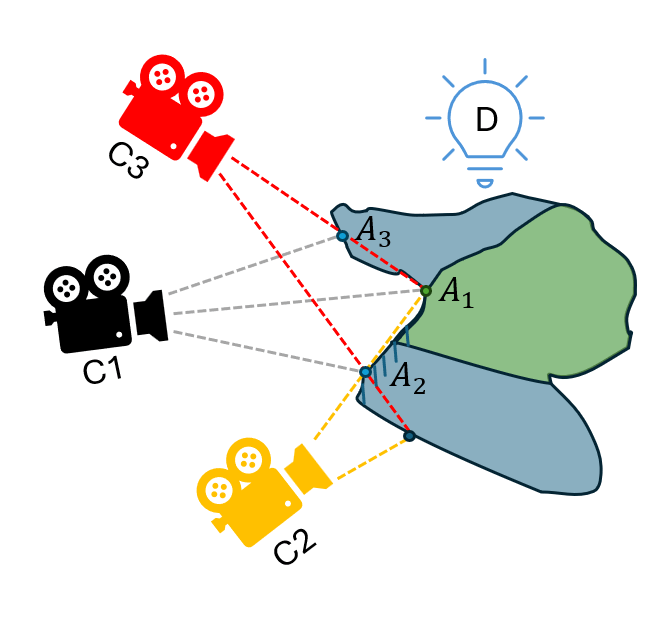}
    \caption{Illustration of how variance can fail as a measure of inconsistency. $C_1$ is the reference, while $C_2$ and $C_3$ are the virtual cameras. Due to occlusion, when warping the rendered images from $C_2$ or $C_3$ to $C_1$, the prediction for $A_1$ will be associated with $A_2$ and $A_3$. However, since $A_2$ and $A_3$ have similar depths with respect to the reference camera and similar RGB values, the depth and color variance among the three observations will be low. 
    On the other hand, if we use differences with the reference depth and color as the criterion, the inconsistency of $A_1$ will be high, which is the correct behavior for a heavily occluded point.
    }
    \label{fig:consistency of vcam}
    \vspace{-10px}
\end{figure}

At the beginning of this stage, each pixel of the reference view is associated with up to $N$ depths and color (RGB-tuples) warped from virtual views.

We proceed by taking the sum of $L_2$ differences between warped color and the color at the pixel of the reference view,
\begin{equation}
    u^{color,r}_i = \frac{\sum^{N}_{v=1}\| c_i^{(r)} - c_i^{(v)} \|}{\sum^{N}_{v=1} \mathbb{1}(d_i^{(v,r)} > 0) }
\end{equation}

where $\mathbb{1}(.)$ indicates the number of valid projections from virtual cameras to the corresponding reference view. We then repeat this process for depth to obtain $u^{depth,r}_i$.
We found that computing the variance of warped depth or color can fail, because virtual views may cluster favoring a depth or color that is inconsistent with the reference view, as shown in Fig.~\ref{fig:consistency of vcam}. Our chosen formulation makes the reference view privileged and allows it to play the central role in determining the uncertainty of its pixels.

Suppose we have $M$ pixels in each reference view, the uncertainty of a given reference pixel is formulated as:
\begin{equation}
u^r_{i} = \sigma(u^{color,r}_i) + \sigma(u^{depth,r}_i)\\
\label{eq:colordepth}
\end{equation}
where $\sigma(.)$ is the softmax function, used to normalize across all reference pixels:
\begin{equation}\label{eq:norm}
    \sigma(z^r_i)
= \frac{e^{z^r_i}}{\sum_{j=1}^{M} e^{z^r_j}}, \qquad z^r_i \in \{u^{color,r}_i,u^{depth,r}_i\}
\end{equation}
Applying the softmax operator to the accumulated differences in color and depth separately normalizes the two types before combining them, which is necessary due to their heterogeneity. The lower bound of both is 0, but their upper bounds are different: color differences can be up to 1.0 per channel, while depth differences depend on the unit of length used in camera extrinsics, viewing distances and the dimensions of the scene. As a result, depth differences have a scene-dependent upper bound, which does not vary too much across images. This is because images of the same scene tend to observe similar maximum depths due to the initialization of the RF using a common point cloud, and because gross outliers in depth fall out of bounds in other images. In the supplement, we show an ablation study comparing the use of color and depth separately and jointly for uncertainty estimation that demonstrates the effectiveness of Eq.~(\ref{eq:colordepth}).

For the results shown in Section~\ref{sec:experiments}, we require at least two valid observations from virtual views to compute the uncertainty of a pixel. (Otherwise, uncertainty is set to its maximum value since the lack of at least two valid observations is a sign of high uncertainty.)

%% file: sec/experiments.tex
We present experiments on multiple datasets, using both NeRF and GS models, specifically Instant NGP \cite{muller2022instant} and 3DGS \cite{kerbl20233dgs}, as backbones. We evaluate the accuracy of uncertainty estimation through \textit{Area Under the Curve (AUC)} analysis of the RGB errors in rendered images. 

\subsection{Datasets}\label{sec:dataset}

\begin{table}[t]
\centering
\small
\caption{Summary of datasets}
\resizebox{\columnwidth}{!}{
\begin{tabular}{l c c}
\hline\hline
\textbf{Dataset} & \textbf{Real / Synthetic} & \textbf{Scene / Capture Setup} \\
\hline
Blender~\cite{mildenhall2020nerf} & Synthetic & Object-centric \\
Local Light Field Fusion(LLFF)~\cite{mildenhall2019llff} & Real & Indoor/Outdoor, forward-facing \\
Light Field(LF)~\cite{yucer2016efficient} & Real & Object-centric \\
Tanks and Temples(TNT)~\cite{knapitsch2017tanks} & Real & Indoor/Outdoor, large-scale \\
Deep
Blending(DB)~\cite{hedman2018deep} & Real & Indoor/Outdoor, hand-held \\
Mip-NeRF 360(M360)~\cite{barron2022mip} & Real & Indoor/Outdoor, object-centric (orbit) \\
\hline\hline
\end{tabular}
}
\vspace{-10pt}
\label{tab:dataset_summary}
\end{table}

We conduct experiments on six synthetic and real datasets, summarized in Table~\ref{tab:dataset_summary}. We evaluated RF models and uncertainty predictors using every eighth image as the test set for real scenes and the standard test split for synthetic scenes. A few-shot setting yields renderings with non-trivial errors and enables meaningful uncertainty estimation. We accomplish this by using 20 images from each scene for training or optimizing the models. In the experiments presented below, we run Instant NGP on LLFF and LF, and 3DGS on M360, Blender, TNT \& DB.

\subsection{Baselines}

We selected widely used, high-performing baselines that span the uncertainty categories reviewed in Section~\ref{sec:related_works}, following Klasson et al.~\cite{klasson2024sources}. For the NeRF backbone (Instant-NGP), we evaluate two approaches: (1) \textbf{Monte Carlo Dropout (MCD)}~\cite{gal2016dropout}, measuring RGB or depth variance over multiple stochastic renderings (drop probability $p=0.2$, $M_{\text{mcd}}=5$ forward passes), and (2) \textbf{Entropy} of the weight distribution of the color samples along each ray~\cite{lee2022uncertainty,kim2022infonerf}. For the 3DGS backbone, we consider three baselines: (1) \textbf{Ensemble}~\cite{klasson2024sources,sunderhauf2023density}, using the variance of rendered RGB images or depth maps across $M_{\text{ens}}=5$ members; each member is trained on a distinct subset of 15 images sampled from the 20 training views; (2) \textbf{FisherRF}~\cite{jiang2024fisherrf}, which quantifies uncertainty via the Hessian matrix of the rendering objective function, which is approximated in way that is better suited to explicit representations; and (3) \textbf{WarpRF}~\cite{safadoust2026warprf}, which was implemented following its authors’ guidelines, and which warps rendered depth maps from all training views to a reference view and computes uncertainty as the average $L_1$ depth discrepancy. We also provide a time complexity analysis of VCURF and each baseline in the supplementary material.

\input{fig/unc_maps}

\subsection{Implementation Details of VCURF}

We used the trained backbones to obtain identical renderings for \methodname and the baselines to enable fair comparisons. 
When using the ensemble, this is not straightforward since each member of the ensemble must render to the same viewpoint before the renderings are averaged to obtain the final one.
As a result, ensembles operate on different renderings with smaller errors, in general.

The parameters of \methodname were set as follows: the number of virtual cameras $N$ is set to 6, the radius of the sphere is set at 10\% of the median depth of the reference view, while the threshold that prevents cameras from being to close to each other is set to 30\% of the radius, i.e. $p_1=0.1$ and $p_2=0.3$. These parameters are kept constant throughout. Ablation studies in the supplement show that VCURF is not sensitive to those hyper-parameters, as long as the sampling radius is smaller than 50\% of the median depth. 

\subsection{Evaluation Metric}

We evaluate uncertainty estimation by measuring the \textit{area under the densification ROC curve (AUC)}~\cite{goli2024bayes,jiang2024fisherrf,klasson2024sources}.  Pixels are sorted according to estimated uncertainties, 
starting with the least uncertain pixels and progressively increasing the density until all pixels are selected. Lower AUC values indicate better performance since they correspond to errors being sorted more precisely.  We also include the AUC for pixels sorted according to RGB error itself (labeled as \textit{Opt.} in the tables) as a lower bound of what is achievable, if all errors were ranked from lowest to highest perfectly.

\input{fig/inst_3dgs}
\input{tab/instantngp}

\input{tab/3dgs}

\subsection{Quantitative and Qualitative Results}\label{sec:results}

In Table~\ref{tab:instantngp}, we provide a quantitative evaluation of uncertainty estimation, comparing AUC scores on the LLFF and LF datasets using InstantNGP~\cite{muller2022instant} as the backbone RF. VCURF outperforms all baselines on average. Entropy often assigns high uncertainty to high-frequency regions, such as the tablecloth in Fig.~\ref{fig:unc} Row 4, even if they are synthesized well. MCD-depth detects errors caused by lack of observations, while MCD-rgb emphasizes intensity changes such as reflections. VCURF is more balanced across all kinds of errors.

In Table~\ref{tab:3dgs}, we provide further AUC evaluations on M360,
Blender, TNT, and DB using 3DGS \cite{kerbl20233dgs} as the RF backbone.
VCURF substantially outperforms FisherRF \cite{jiang2024fisherrf} and
performs comparably to ensembles despite its lower training and storage costs.
This may reflect the local validity of FisherRF's Laplace approximation,
whereas VCURF directly measures finite rendering variations. VCURF also
outperforms WarpRF despite their similarity. We attribute this to the fact that the virtual cameras are configured identically around each reference camera making consistency checks directly comparable, as opposed to WarpRF that relies on the distribution of input cameras.
When VCURF and Ensemble start with similar errors on M360 (see Opt. vs. Ensemble Opt.), VCURF predicts uncertainty more precisely. Ensemble-rgb is the best predictor on Blender.

\subsection{Application: Floater Removal}
\begin{figure}[t]
    \centering
    \vspace{-15pt}
    \includegraphics[width=0.95\columnwidth]{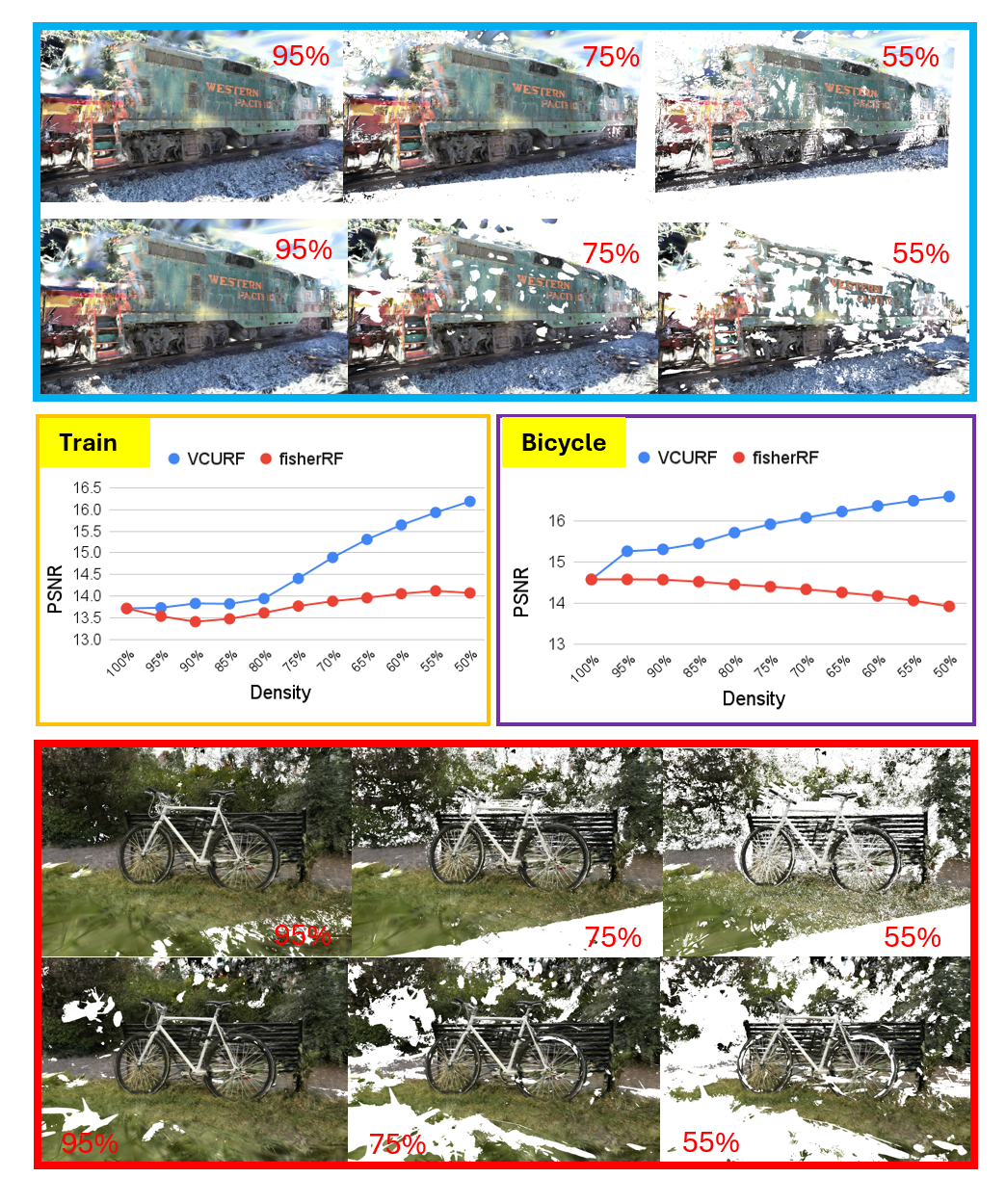}
    \vspace{-13pt}
    \caption{Floater clean-up for \textit{Train (T\&T)} and \textit{Bicycle (M360)}. For each group, the top images are filtered by VCURF while the bottom ones by FisherRF. The plots show PSNR after progressive noise removal.}
    \label{fig:floater_remover}
    \vspace{-15pt}
\end{figure}

We present floater removal as an application in Fig.~\ref{fig:floater_remover}, following a setting similar to Bayes' Rays (Sec.~5.2 and Fig.~8 in \cite{goli2024bayes}). This experiment evaluates how well pixel-wise uncertainty identifies rendering artifacts. Given the same pretrained 3DGS model and rendered images, we estimate pixel-wise uncertainty using FisherRF and our method, respectively. We then progressively remove pixels with the highest uncertainty and evaluate the PSNR of the remaining pixels. If the estimated uncertainty is well aligned with rendering errors, removing high-uncertainty pixels should progressively increase PSNR. As shown in Fig.~\ref{fig:floater_remover}, our method yields a substantially larger PSNR improvement than FisherRF as more uncertain pixels are removed, indicating that its uncertainty estimates more accurately localize floaters and other rendering errors.

%% file: fig/unc_maps.tex
\begin{figure*}[t]
\centering
\begin{tabular}{ccccc}
\includegraphics[width=0.16\textwidth]{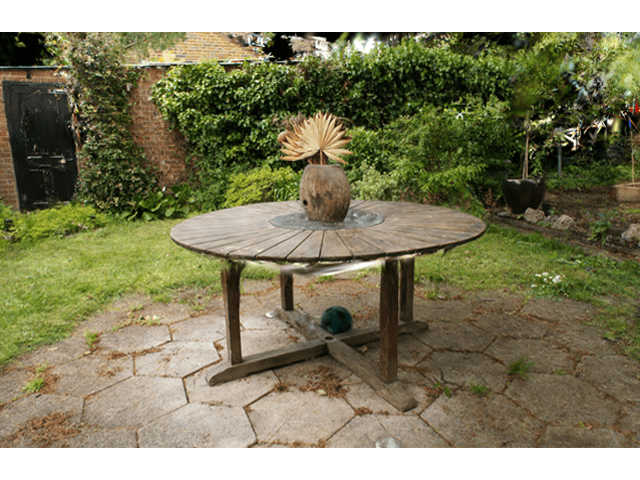} & 
\includegraphics[width=0.16\textwidth]{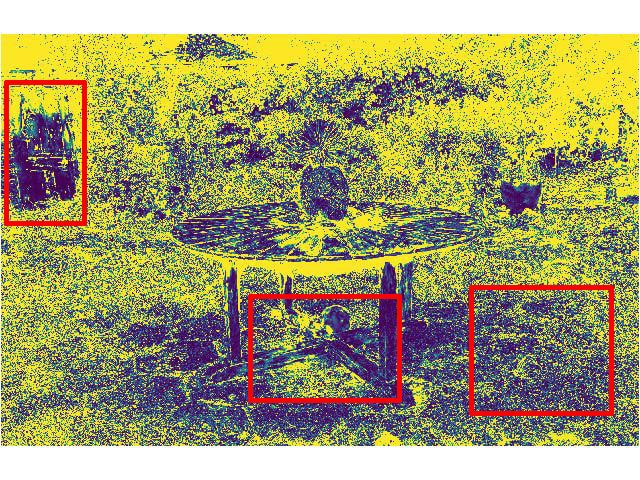} & 
\includegraphics[width=0.16\textwidth]{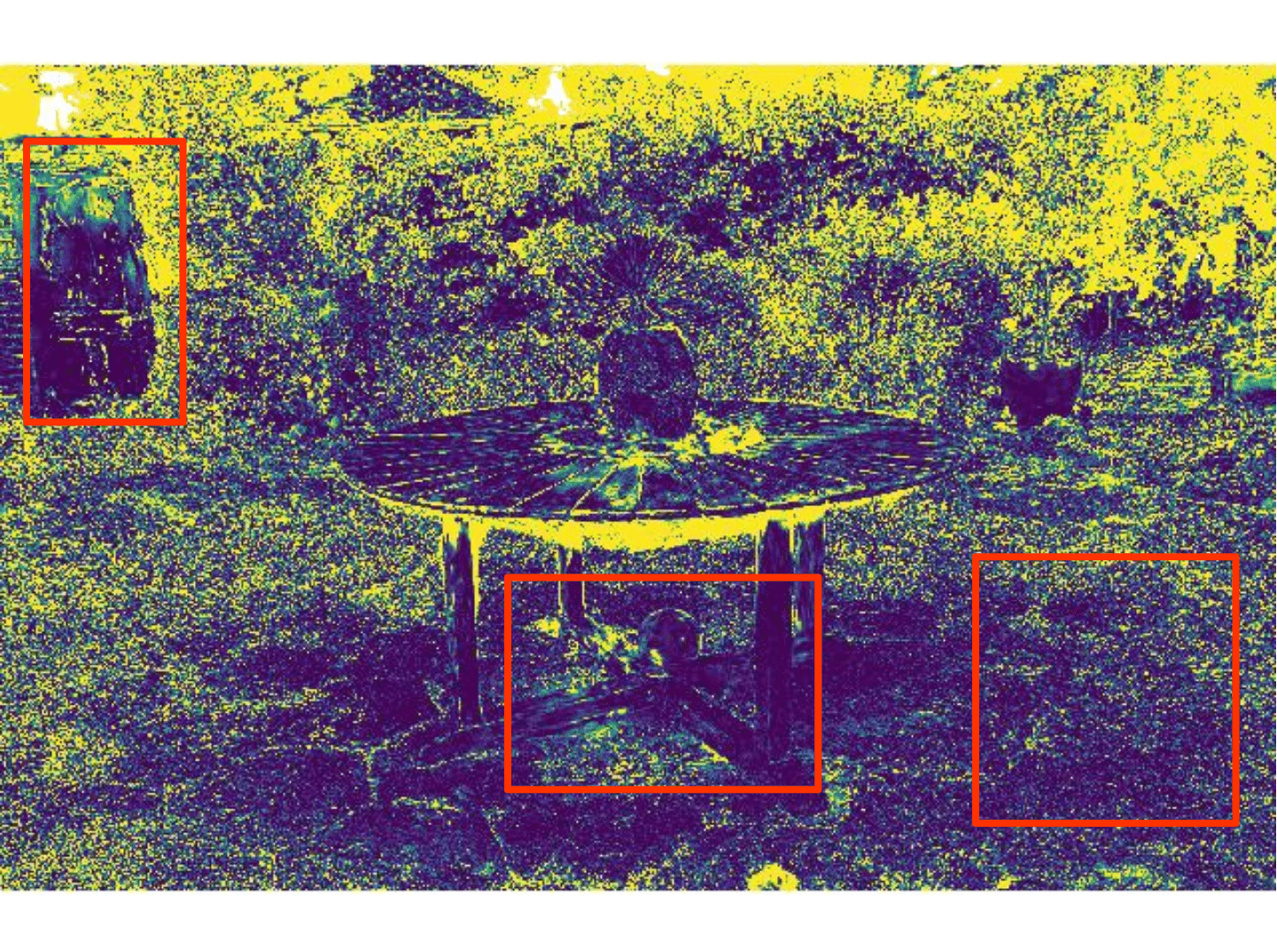} & 
\includegraphics[width=0.16\textwidth]{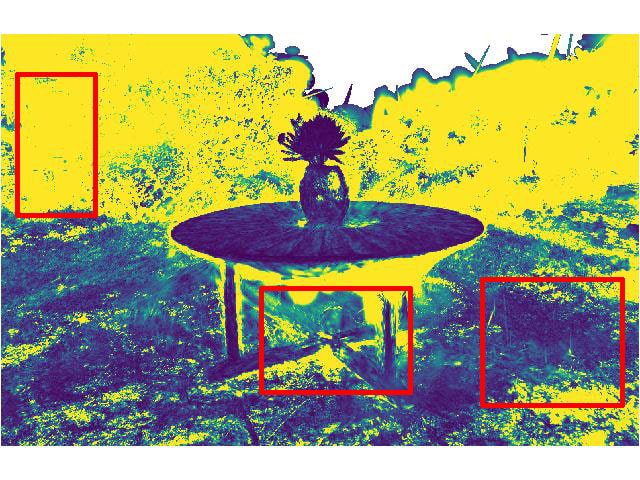} & 
\includegraphics[width=0.16\textwidth]{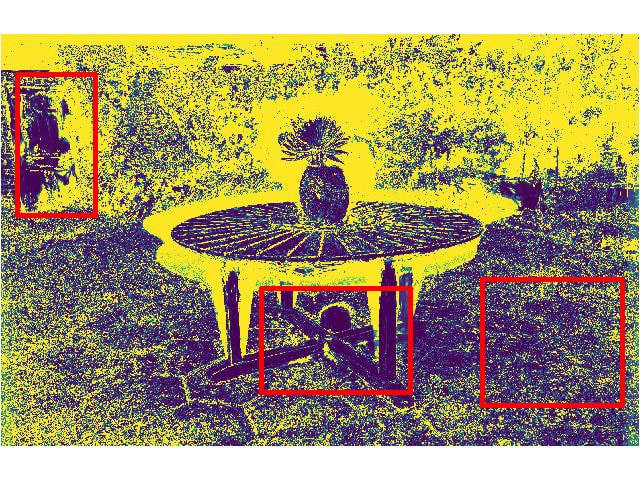} \\
3DGS & RGB error & WarpRF & Ensemble-depth & VCURF \\

\includegraphics[width=0.16\textwidth]{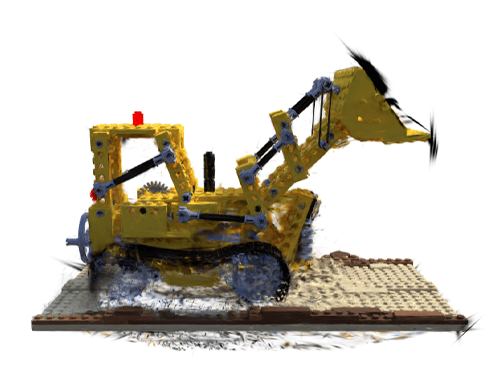} & 
\includegraphics[width=0.16\textwidth]{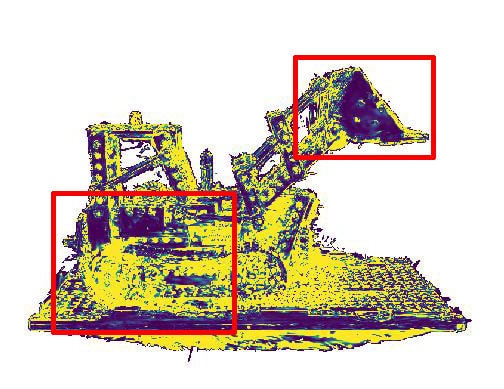} & 
\includegraphics[width=0.16\textwidth]{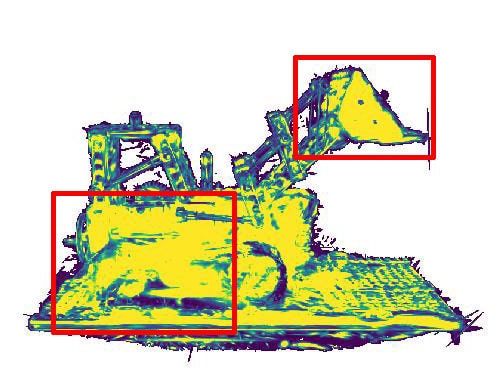} & 
\includegraphics[width=0.16\textwidth]{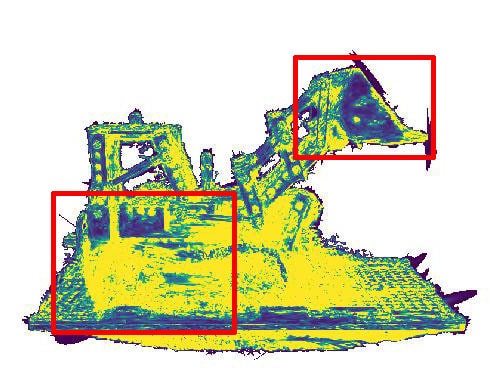} & 
\includegraphics[width=0.16\textwidth]{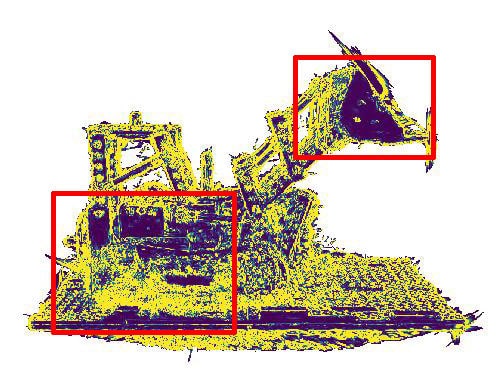} \\
3DGS & RGB error & FisherRF & Ensemble-rgb & VCURF \\

\includegraphics[width=0.16\textwidth]{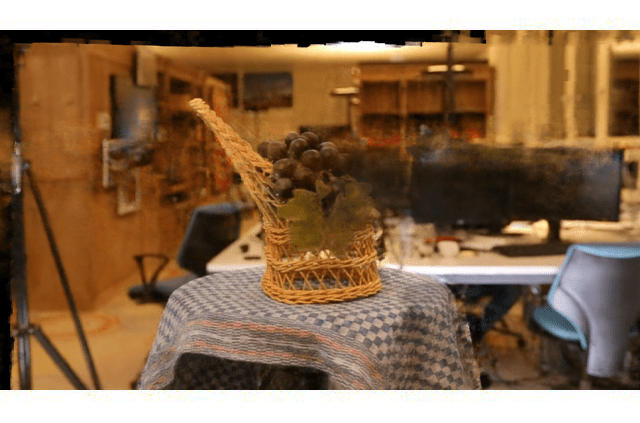} & 
\includegraphics[width=0.16\textwidth]{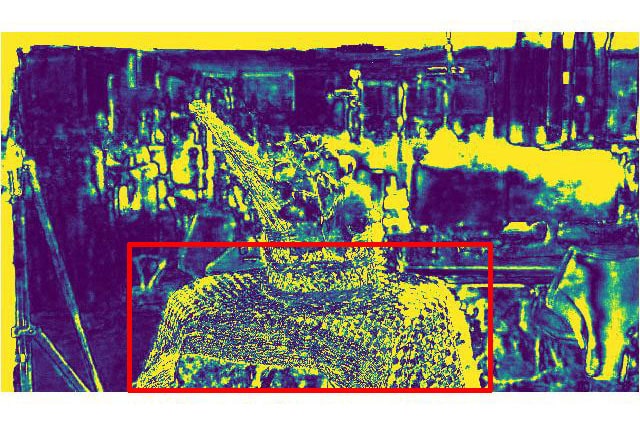} & 
\includegraphics[width=0.16\textwidth]{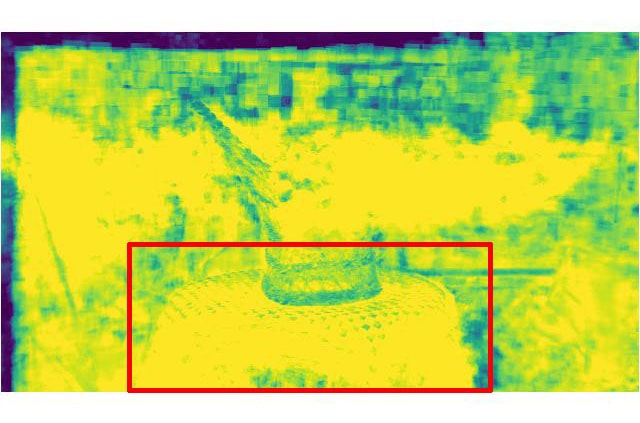} & 
\includegraphics[width=0.16\textwidth]{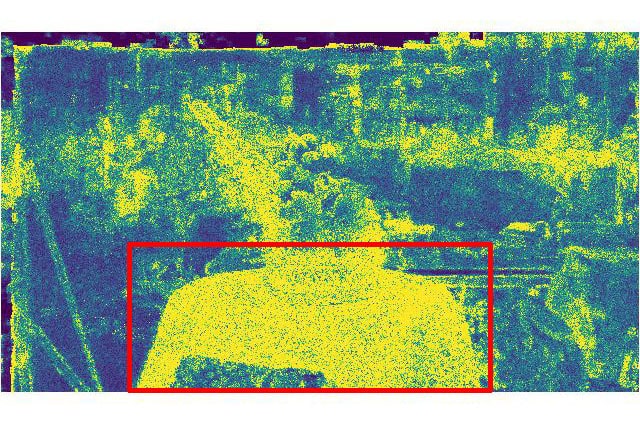} & 
\includegraphics[width=0.16\textwidth]{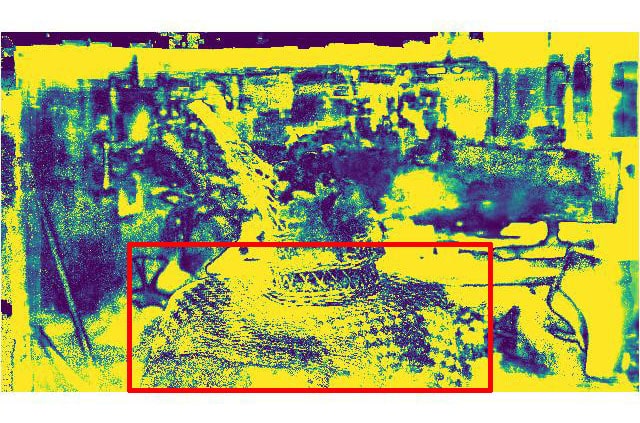} \\
Instant-NGP & RGB error & Entropy & MCD-rgb & VCURF \\

\includegraphics[width=0.16\textwidth]{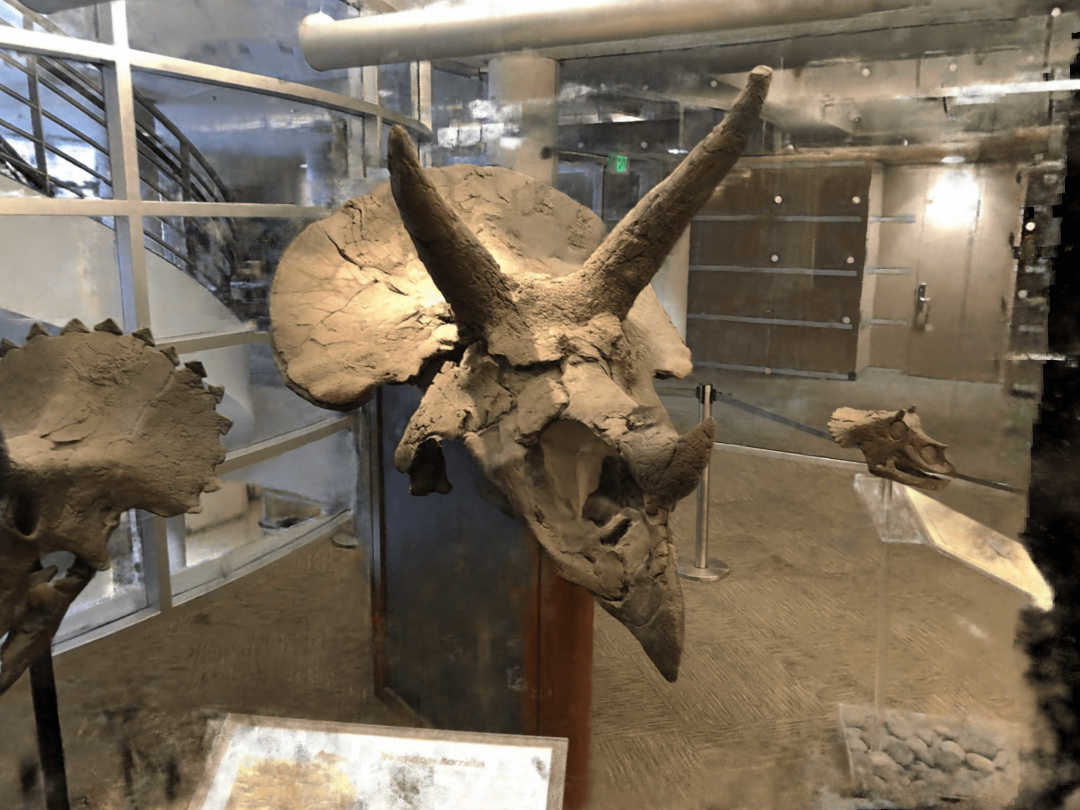} & 
\includegraphics[width=0.16\textwidth]{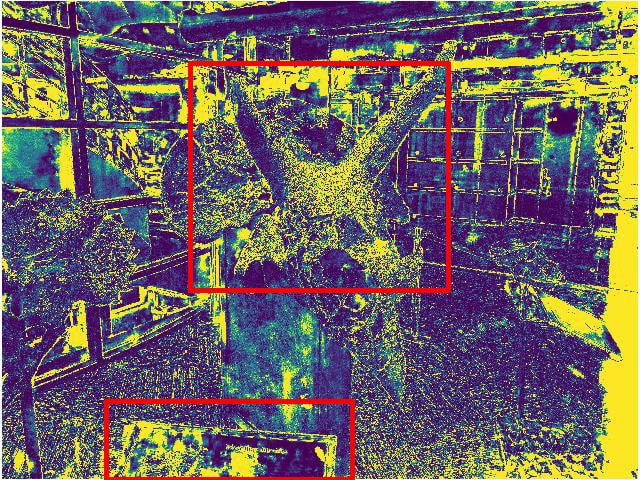} & 
\includegraphics[width=0.16\textwidth]{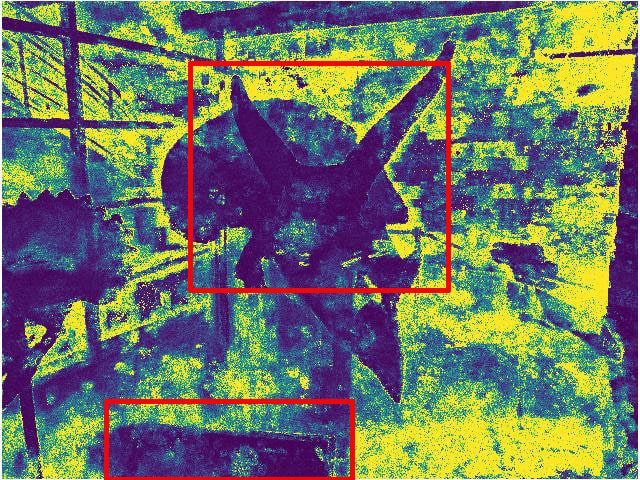} & 
\includegraphics[width=0.16\textwidth]{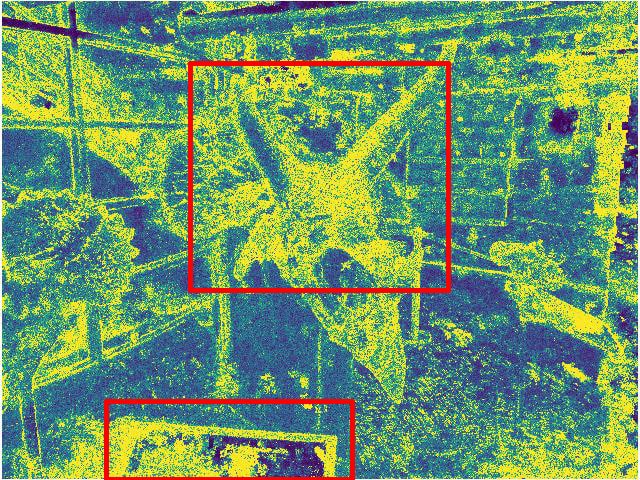} & 
\includegraphics[width=0.16\textwidth]{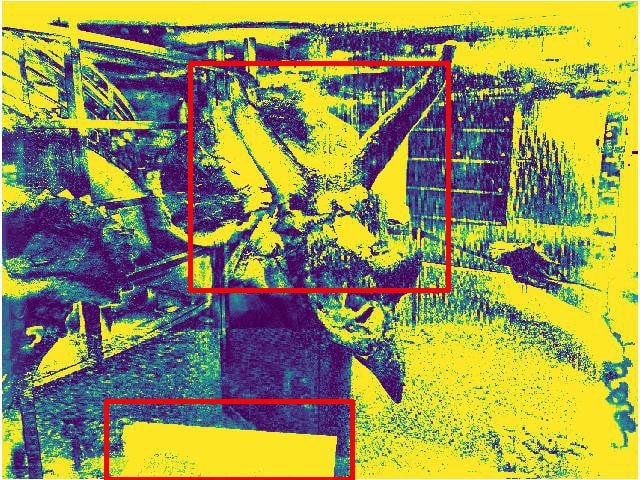} \\
Instant-NGP & RGB error & MCD-depth & MCD-rgb & VCURF \\

\end{tabular}
\caption{Rendered images, error maps, and corresponding uncertainty maps for one representative scene from each dataset. Within each row, results share the same backbone (indicated in the first column). In these scenes, the RF models can generate reasonably accurate novel views but still exhibits visible errors. We use the color map of Fig.~\ref{fig:overview} where yellow represents high and blue represents low values.
The uncertainty maps of VCURF have better alignment with the error maps compared to the baselines, especially in the red-highlighted regions. The datasets from top to bottom are as follows: [M360, Blender, LF, LLFF]}
\vspace{-10pt}
\label{fig:unc}
\end{figure*}

%% file: tab/instantngp.tex
\begin{table}[b]
\caption{Comparison of AUC scores on the LLFF and LF datasets using InstantNGP~\cite{muller2022instant} as the backbone RF.}
\centering
\resizebox{0.8\columnwidth}{!}{
\begin{tabular}{lccccc}
\hline\hline
\textbf{LLFF}   & room              & t-rex             & fortress          & horns             & avg.              \\ \hline
Opt.            & 0.0423            & 0.0622            & 0.0287            & 0.0967            & 0.0575            \\
MCD-depth       & \textbf{0.1897}   & 0.3898            & 0.2455            & 1.2493            & 0.5186            \\
MCD-rgb         & 0.3533            & \textbf{0.3094}   & \textbf{0.0822}   & 0.7804            & 0.3813            \\
Entropy         & 0.5001            & 0.8343            & 0.3667            & 2.0371            & 0.9346            \\ 
VCURF           & 0.2050            & 0.4116            & 0.1783            & \textbf{0.6776}   & \textbf{0.3681}   \\\hline\hline

\textbf{LF}     & africa            & basket            & statue            & torch             & avg.              \\ \hline
Opt.            & 0.2532            & 0.9589            & 0.3086            & 0.3094            & 0.4575            \\
MCD-depth       & 1.5158            & 6.1896            & \textbf{1.0453}   & \textbf{1.3141}   & 2.5162            \\
MCD-rgb         & \textbf{1.0044}   & 4.9930            & 1.3709            & 2.2596            & 2.4070            \\
Entropy         & 2.1657            & 6.1266            & 2.3356            & 2.2596            & 3.2219            \\ 
VCURF           & 1.2037            & \textbf{4.2323}   & 1.4525            & 1.6421            & \textbf{2.1327}   \\\hline\hline
\end{tabular}
}
\label{tab:instantngp}
\end{table}

%% file: tab/3dgs.tex
\begin{table}[t]
\centering
\caption{Comparison of AUC scores on mip-NeRF 360, Blender, Tanks and Temples, and Deep Blending datasets using 3DGS~\cite{kerbl20233dgs} as the backbone RF.}
\label{tab:3dgs}
\resizebox{\columnwidth}{!}{
\begin{tabular}{lccccc}
\hline\hline
\textbf{M360} &  kitchen &  garden &  bicycle &  counter &  avg. \\
\hline
    Opt.              & 0.0837 & 0.1040 & 0.2158 & 0.0902 & 0.1234 \\
    FisherRF           & 0.8291 & 1.1571 & 2.0789 & 1.2301 & 1.3238 \\
    WarpRF             & 0.4902 & 0.5268 & 1.3243 & 0.7166 & 0.7645 \\
    VCURF              & \textbf{0.4837} & \textbf{0.4816} & \textbf{1.2062} & \textbf{0.4945} &  \textbf{0.6665} \\
\hline
    Ensemble Opt      & 0.0497 & 0.1057 & 0.1682 & 0.0780 & 0.1004 \\
    Ensemble - rgb     & \textbf{0.1422} & 1.1441 & \textbf{1.0737} & \textbf{0.3951} &  \textbf{0.6888} \\
    Ensemble - depth   & 0.1811 & \textbf{0.9296} & 1.3921 & 0.5325 & 0.7588 \\ 
\hline\hline

\textbf{Blender} &  ship &  lego &  chair &  drums &  avg. \\
\hline
    Opt.               & 0.0838 & 0.0735 & 0.1029 & 0.1783 & 0.1096 \\
    FisherRF           & 0.4871 & 0.8420 & 0.9659 & \textbf{1.2901}& 0.8963 \\
    VCURF              & \textbf{0.3595} & \textbf{0.5420} & \textbf{0.5732} & 1.3267 &   \textbf{0.7004} \\
\hline
    Ensemble Opt.      & 0.0382 & 0.0534 & 0.0797 & 0.1033 & 0.0686 \\
    Ensemble-rgb       & \textbf{0.0990} & \textbf{0.2648} & \textbf{0.2437} & \textbf{0.3100} &  \textbf{0.2294} \\
    Ensemble-depth     & 0.2773 & 0.3757 & 0.4191 & 1.9996 & 0.7679 \\
\hline\hline

\textbf{TNT} \& \textbf{DB} &  train &  trunk &  drjohnson &  playroom &  avg. \\
\hline
    Opt.               & 1.0202 & 0.1632 & 0.4160 & 0.3597 & 0.4898 \\
    FisherRF           & 4.5159 & 1.7844 & 2.9768 & 3.7496 & 3.2567 \\
    WarpRF             & 4.0230 & 0.9487 & 1.6724 & \textbf{1.3178} & 1.9905 \\
    VCURF              & \textbf{3.7451} & \textbf{0.9036} & \textbf{1.6655} & 1.5269 & \textbf{1.9603} \\
\hline
    Ensemble Opt.      & 0.8885 & 0.1330 & 0.3560 & 0.3350 & 0.4281 \\
    Ensemble-rgb       & \textbf{2.8199} & \textbf{0.4478} & \textbf{1.7186} & \textbf{1.6595} & \textbf{1.6615} \\
    Ensemble-depth     & 3.3107 & 0.6733 & 1.8607 & 1.9222 & 1.9417 \\
\hline\hline

\end{tabular}
}
\vspace{-10pt}
\end{table}

%% file: sec/discussion.tex
Most uncertainty-based view-selection methods estimate pixel-wise
uncertainty and aggregate it into a view-level score, typically by averaging 
the uncertainty over all pixels of each candidate view. They then select the
view with the highest score, implicitly assuming that:
(i) uncertainty approximates the rendering error, and (ii) larger rendering error implies greater information gain when the view is added to the training set. Our experiments support the first assumption: VCURF accurately identifies pixels with large rendering errors. In this section, we examine whether the second assumption necessarily follows.

\paragraph{Rendering error is not equivalent to view utility.}
We conduct a diagnostic experiment following the Batch Active View Selection protocol of FisherRF \cite{jiang2024fisherrf}. We compare VCURF with an \emph{error oracle} that has access to the true pixel-wise RGB rendering error of each candidate view. Candidates are ranked by average pixel-wise uncertainty for VCURF and average true rendering error for the oracle. For VCURF, the terms in Eq.~(\ref{eq:norm}) are normalized jointly over all candidate-view pixels.

If larger rendering error implied greater information gain, and as a result larger view utility, the error oracle would outperform its uncertainty-based approximation. However, Table~\ref{tab:view_sel} shows the opposite trend: view selection using VCURF achieves better average PSNR, SSIM, and LPIPS than selection using the true rendering error. The oracle performs slightly better for only a few scene--metric combinations.

\input{tab/view_sel}

This result does not mean that VCURF estimates rendering error more accurately than the oracle; by definition, it cannot. Instead, it shows that accurately ranking candidate views according to their current rendering errors is not equivalent to ranking them according to their value for subsequent RF optimization. A candidate view with large error identifies where the current model struggles, but does not reveal how much novel and non-redundant supervision that view contributes relative to the existing training set. Consequently, improving rendering-error estimation does not necessarily yield a corresponding improvement in active view selection.

A useful interpretation is to formulate view selection as a continual learning problem, in which the RF must synthesize novel views from a sequence of training sets that grow as observations are acquired. Based on the analysis of Ramesh and Chaudhari \cite{ramesh2021model}, an RF can be improved by: (i) reducing empirical error on the current training views, which is typically already small; (ii) reducing excess error on novel views, i.e. 
the gap between the model's
population error and the lowest error achievable within its hypothesis class; and (iii) acquiring informative new samples for training. Pixel-wise uncertainty primarily identifies (ii), while contributing to (iii) only indirectly. In particular, a high-error candidate may overlap substantially with previously acquired views and therefore provide limited additional supervision. Repeatedly selecting such views can yield diminishing returns even when their rendering errors are estimated perfectly.

PRIMU \cite{gottwald2026primu} provides further evidence under the FisherRF protocol: training-view coverage outperforms methods based on uncertainty or approximate information gain. By prioritizing poorly covered candidates, PRIMU targets novel observations rather than regions of high current error. Together with our oracle experiment, this indicates that local rendering difficulty and global acquisition utility are distinct.

Uncertainty- and coverage-based criteria nevertheless provide complementary information. Pixel-wise uncertainty captures local, scene-dependent reconstruction difficulty associated with occlusions, concavities, and spatially varying appearance or geometric detail. Global criteria based on geometric coverage \cite{kopanas2023improving} or camera distribution and redundancy \cite{xiao2024nerfdirector,wang2024redundancy} more directly encourage novel observations, but may overlook these image-dependent challenges. An effective active-reconstruction strategy should therefore combine local uncertainty with a global estimate of observation novelty, coverage, and redundancy.

Our diagnostic experiment is intended to expose the limitation of treating rendering error as a complete acquisition objective, rather than to establish VCURF as a complete view-selection criterion. Developing and systematically evaluating a global acquisition strategy that integrates uncertainty reduction with novel and non-redundant coverage is an exciting direction for future work.

%% file: tab/view_sel.tex
\begin{table}[t]
\caption{Diagnostic active-view-selection experiment comparing candidate
ranking by VCURF uncertainty with ranking by true average rendering error.
Despite having perfect access to the current rendering errors, the error-based
oracle produces worse average NVS performance, showing that rendering error is
not equivalent to view utility.}
\label{tab:view_sel}
\centering
\resizebox{0.75\columnwidth}{!}{
\begin{tabular}{lccccc}
\hline\hline
 &
garden &
bicycle &
room &
stump &
avg. \\
\hline

\textbf{\underline{PSNR}} &&&&&\\
Oracle       & \textbf{21.72} & 16.46 & \textbf{23.02} & 19.51 & 20.18 \\
VCURF             & 21.40 & \textbf{17.89} & 22.39 & \textbf{19.66} & \textbf{20.38} \\
\hline
\textbf{\underline{SSIM}} &&&&&\\
Oracle       & 0.634 & 0.353 & 0.788 & 0.435 & 0.553 \\
VCURF             & \textbf{0.652} & \textbf{0.425} & \textbf{0.793} & \textbf{0.487} & \textbf{0.589} \\
\hline
\textbf{\underline{LPIPS}} &&&&&\\
Oracle       & 0.276 & 0.491 & 0.317 & 0.460 & 0.386 \\
VCURF             & \textbf{0.267} & \textbf{0.413} & \textbf{0.312} & \textbf{0.427} & \textbf{0.355} \\
\hline\hline
\end{tabular}
}
\vspace{-10pt}
\end{table}

%% file: sec/conclusion.tex
We have presented \methodname, a new approach for uncertainty prediction that is applicable to both NeRF and GS models, is simple to use, and has low sensitivity to hyperparameter settings. By measuring consistency across locally sampled virtual views, \methodname treats the RF as a black box, requires no model modification or additional training, and incurs only the cost of additional rendering. Quantitative and qualitative experiments show that it predicts rendering errors more accurately than existing methods, including FisherRF, while remaining robust to hyperparameter settings and making few assumptions about the underlying representation.

Our diagnostic view-selection experiment  shows that accurate rendering error does not necessarily imply high acquisition utility: uncertainty identifies where the model struggles but does not capture global coverage or observation redundancy. Future work will therefore combine local uncertainty with global coverage for active reconstruction and distinguish uncertainty sources, including epistemic and aleatoric uncertainty as well as errors caused by occlusion and camera miscalibration.

%% file: sec/X_suppl.tex
\clearpage
\maketitlesupplementary

\renewcommand\thefigure{S.\arabic{figure}}  
\renewcommand\thetable{S.\arabic{table}}
\renewcommand\thesection{S.\arabic{section}}

\setcounter{page}{1}
\setcounter{section}{0}
\setcounter{table}{0}
\setcounter{figure}{0}

In this document, we present ablation studies (Section~\ref{sec:ablation}), a storage and complexity analysis for VCURF and competing methods (Section~\ref{sec:time_complx}), and additional qualitative results (Section~\ref{sec:unc_map}).

\section{Ablation Studies}\label{sec:ablation}
\input{sup/ablation}

\section{Storage and Time Complexity Analysis}
\label{sec:time_complx}
\input{sup/time_complexity}

\section{Qualitative Results}
\label{sec:unc_map}

\input{sup/qualitative}

We provide additional qualitative results to complement Section 4 of the main paper. For better visualization, please zoom in to observe the detailed differences in each image. Specifically, we include one additional scene from each dataset. For these scenes, we present the following comparisons:
(1) Rendered images, (2) Error maps, (3) Uncertainty maps. Results  for various baselines are shown alongside our proposed method, VCURF.

%% file: sup/ablation.tex
We performed ablation studies on the hyper-parameters of VCURF shown in Section 3.1 in the main paper. Specifically, we calculate AUC values when running VCURF with different parameters: the radius of the sphere $r_s$ within which cameras are sampled, the number of virtual cameras $N$, and different components of the uncertainty estimate. We also show the standard deviation (std.) of the AUC values across different settings on the same scene. Lower std. is desirable, as it indicates that our method is less sensitive to that hyper-parameter. We conduct another ablation study evaluating randomly sampling virtual camera positions versus manual specification. We run all the ablation studies on 3 scenes with 3DGS, i.e. \textit{ship} from Blender, \textit{kitchen} from Mip-Nerf 360$^\circ$ and \textit{playroom} from Tank and Temples. 

\subsection{Sampling Radius}
\label{sec:r_scale}
The sampling radius can be viewed as the upper bound of baselines between cameras. We first set the number of virtual cameras to  6 and run VCURF with different sampling radii. We show the AUC value of different sampling radii in Table~\ref{tab:r_scale}, the small standard deviation indicates that VCURF is not sensitive to this hyper-parameter.
\input{tab/r_scale}

\subsection{Number of Virtual Cameras}
\label{sec:n_vcam}
\begin{figure}[t]
    \centering
    \vspace{-8pt}
    \resizebox{0.8\columnwidth}{!}{\includegraphics[width=0.9\columnwidth]{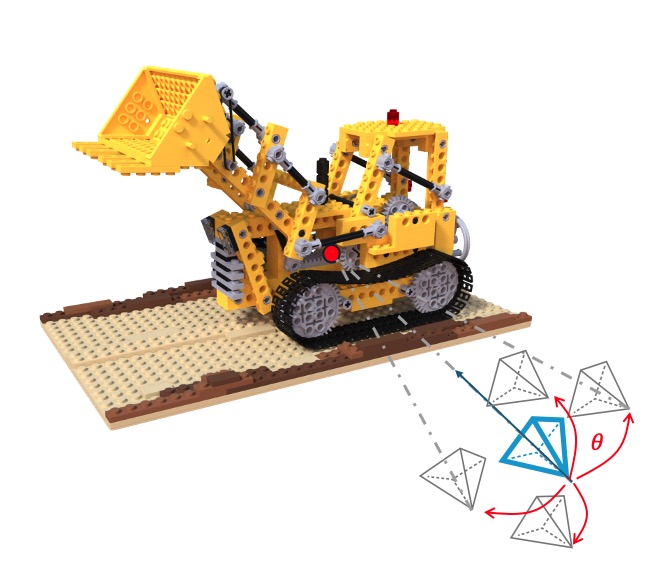}
    }\vspace{-10pt}
    \caption{Illustration of fixed camera configuration. Each virtual camera (gray) is rotated around the scene center $O$ by a small angle $\theta$ along either $X$ or $Y$ axis relative to the reference camera (blue). Consequently, the baselines and distances from the virtual cameras to the scene center remain identical.}
    \label{fig:fixed}
\vspace{-20pt}
\end{figure}

We then set the sampling radius to be $0.1 d_{med}$ and run VCURF with different numbers of virtual cameras. Here, we also compare the random sampling configuration (as we describe in Fig.~1 (a) and Section 3.1) vs. a fixed configuration (illustrated in Fig.~\ref{fig:fixed}). We only compute the standard deviation of the random configurations. Based on the results in Table~\ref{tab:n_vcam}, we found that: (1)~VCURF is not sensitive to the number of virtual cameras, since the standard deviation is very small. (2)~Sampling more than two virtual views is better, since it decreases the likelihood of only estimating in a poor-quality virtual view. (3)~A random configuration is preferable to a fixed configuration, as it promotes variability in baselines and distances to the scene center among the virtual views. Please also see a qualitative example in Fig.~\ref{fig:ship_roc}.

\input{tab/n_vcam}

\input{sup/ship_roc}

\subsection{Components of the Uncertainty Estimate}
\label{sec:comp}
We also conduct an ablation study on how different components affect the performance of VCURF.  We compare variants of VCURF considering the L2 differences of depth or RGB between the reference and virtual cameras separately with the complete algorithm that considers both.  We present the results in Table~\ref{tab:comp}, highlighting the necessity of combining both RGB and depth features rather than using them in isolation.
\input{tab/comp}

%% file: tab/r_scale.tex
\begin{table}[b]
\centering
\caption{AUC values for different sampling radii, expresssed as fractions of the median depth $d_{med}$, and standard deviation across all runs}
\resizebox{0.9\columnwidth}{!}{
\begin{tabular}{ccccc}
\hline\hline
    \textbf{Sample radius} & ship   & kitchen & playroom & avg.   \\ \hline
0.05$d_{med}$ & 0.3606 & 0.5407  & 1.5623   & 0.8212 \\
0.1$d_{med}$ & 0.3565 & 0.4837  & 1.5269   & \textbf{0.7890} \\
0.3$d_{med}$  & 0.3565 & 0.6031  & 1.7759   & 0.9118 \\
0.5$d_{med}$  & 0.3543 & 0.6387  & 1.9711   & 0.9880 \\ \hline
 std.  & 0.0026 & 0.0685  & 0.2064   & 0.0902 \\
\hline\hline
\end{tabular}}
\vspace{-5pt}
\label{tab:r_scale}
\end{table}

%% file: tab/n_vcam.tex
\begin{table}[b]
\centering
\caption{AUC scores for different number of virtual cameras}
\resizebox{0.9\columnwidth}{!}{
\begin{tabular}{ccccc}
\hline
\hline
\textbf{$N$ vcam} & ship   & kitchen & playroom & avg.   \\ \hline
4 (fixed)            & 0.3760 & 0.5744  & 1.5779   & 0.8428 \\ \hline
2                    & 0.3594 & 0.6699  & 1.9731   & 1.0008 \\
4                    & 0.3566 & 0.5762  & 1.5654   & 0.8327 \\
6                    & 0.3565 & 0.4837  & 1.5269   & \textbf{0.7890} \\
8                    & 0.3607 & 0.5737  & 1.6784   & 0.8709 \\
10                   & 0.3627 & 0.5221  & 1.5251   & 0.8033 \\ 
\hline
std.                  & 0.0027 & 0.0701  & 0.1891   & 0.0851 \\
\hline
\hline
\end{tabular}}
\vspace{-5pt}
\label{tab:n_vcam}
\end{table}

%% file: sup/ship_roc.tex
\begin{figure}[tb]
\vspace{-5pt}
    \centering
    \begin{tabular}{cc}

\includegraphics[width=0.45\columnwidth ]{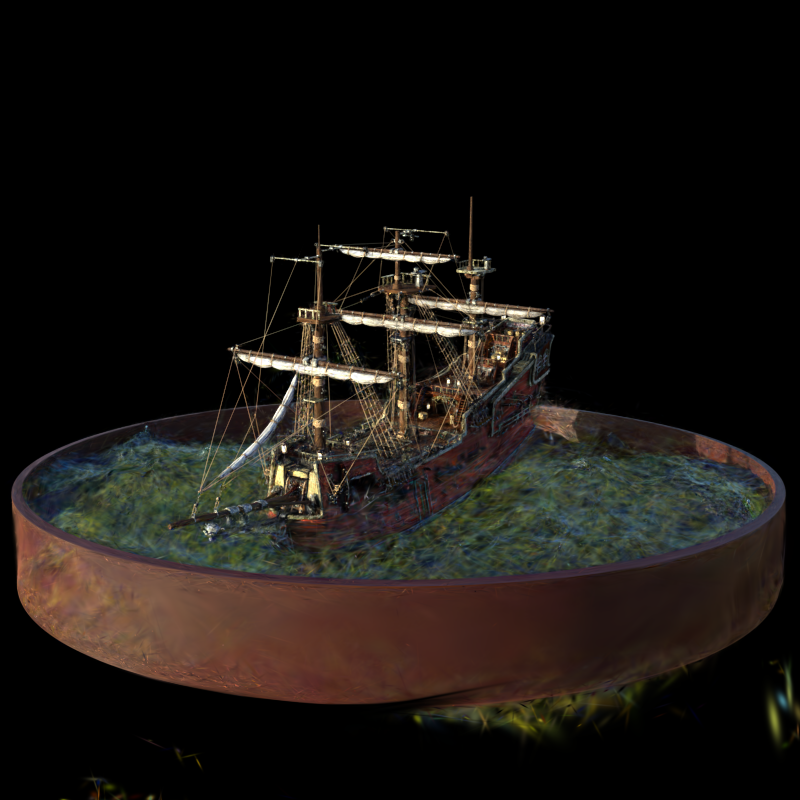} &
\includegraphics[width=0.45\columnwidth ]{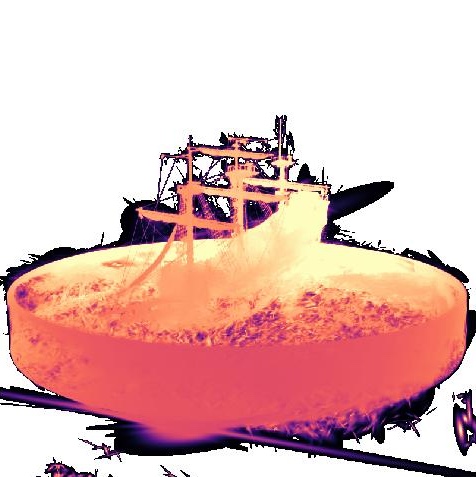}
 \\

Rendered image & Depth
 \\
      \includegraphics[width=0.45\columnwidth ]{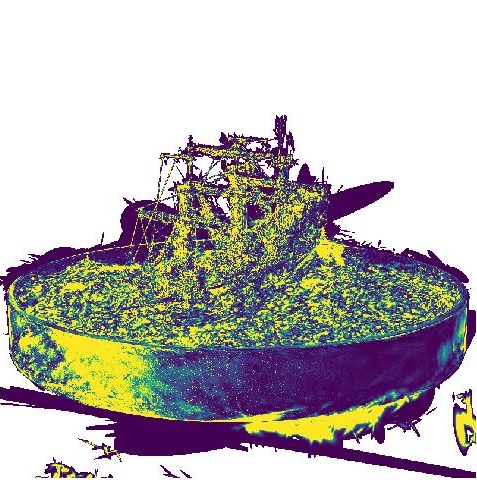}   & \includegraphics[width=0.45\columnwidth ]{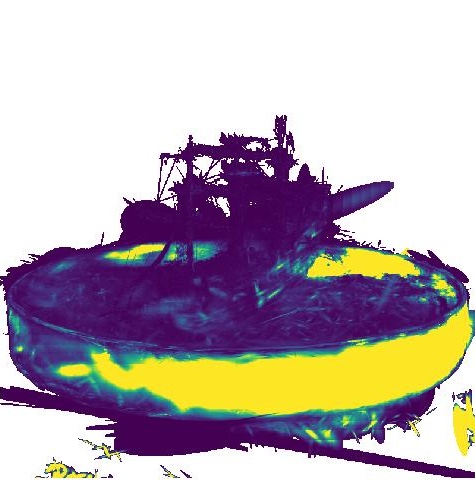} \\
        RGB error & FisherRF  \\

\includegraphics[width=0.45\columnwidth]{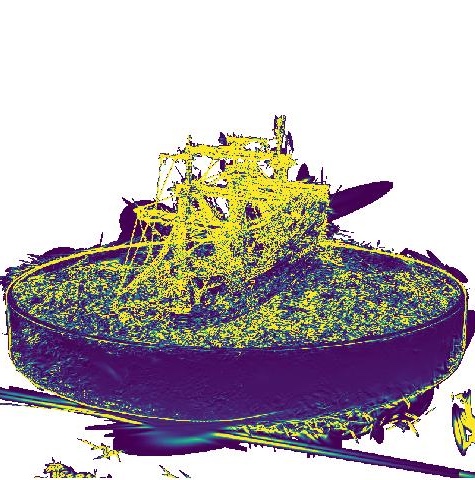}  &
       \includegraphics[width=0.45\columnwidth ]{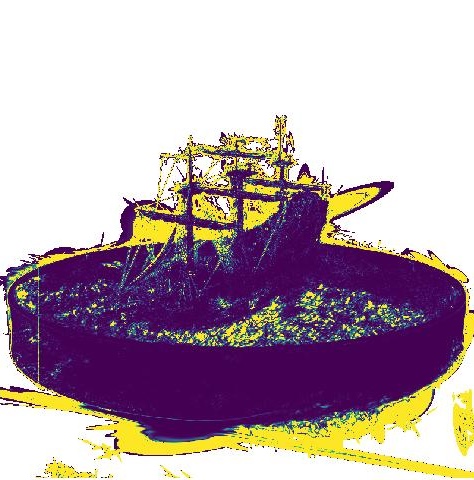}   \\
       VCURF - random & VCURF - fixed   \\

       \multicolumn{2}{c}{
\includegraphics[width=0.9\columnwidth ]{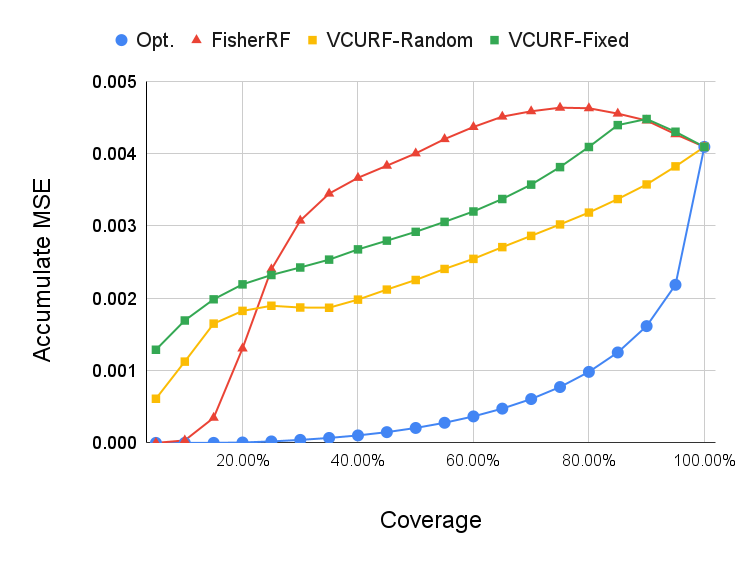}

} \\
          
    \end{tabular} 
    \vspace{-16pt}
    \caption{Rendered RGB, depth, error maps using 3DGS and uncertainty maps;  and densification curves for \textit{ship} from Blender. (Yellow represents high and blue represents low values.) We only set 4 virtual cameras in each configuration. VCURF with random configuration is better than the fixed version and FisherRF. As shown in the uncertainty map, it can successfully highlight the areas of the ship that cause more errors due to the complexity of the structure.}
\label{fig:ship_roc}
\end{figure}

%% file: tab/comp.tex
\begin{table}[htb]
\centering
\caption{AUC scores for different components of VCURF.} 
\resizebox{0.9\columnwidth}{!}{
\begin{tabular}{ccccc}
\hline
\hline
\textbf{Components} & ship   & kitchen & playroom & avg.   \\ \hline
depth-only              & 0.5162 & 0.8291  & 3.7496   & 1.6983 \\
rgb-only            & 0.3033 & 0.8366  & 3.5968   & 1.5789 \\
depth+rgb            & 0.3565 & 0.4837  & 1.5269   & \textbf{0.7890} \\ \hline
\hline
\end{tabular}}
\vspace{-5pt}
\label{tab:comp}
\end{table}

%% file: sup/time_complexity.tex
We analyze the storage and time complexity of computing a pixel-wise uncertainty map for a single candidate view. Let $r$ denote image resolution, $n$ the RF parameter count, $k$ the maximum samples per ray/tile in volumetric rendering, and $M$ the number of MC/ensemble passes (also the number of virtual cameras in VCURF), with $M \ll k \ll n$ in practice.

\textbf{Storage.} All methods store the base RF ($O(n)$). \textbf{Ensemble} must store $M$ models ($O(Mn)$), \textbf{FisherRF} requires additional $O(n)$ space for the Hessian, whereas VCURF computes uncertainty on-the-fly without additional storage requirements.

\textbf{VCURF vs. MCD.} Both require $M$ renderings, giving $O(Mr)$ per view; MCD additionally needs dropout-enabled training. 

\textbf{VCURF vs. Ensemble.} Ensemble renders $M$ independently trained RFs, resulting in $O(Mr)$ per-view inference time and $O(Mn)$ storage.

\textbf{VCURF vs. Entropy.} Entropy computes per-ray statistics via additional ray marching, yielding $O(kr)$ runtime (typically $> O(Mr)$).

\textbf{VCURF vs. FisherRF.} FisherRF uses a model-dependent second-order (Hessian/Fisher) approximation built from parameter-derivative computations, leading to a cost of O(n) with a large constant; it can reuse pre-computed derivatives across views, if the model is not updated.

\textbf{VCURF vs. WarpRF.} WarpRF, which is also black-box, pre-renders (and warps from) \textit{all} training views, so its upfront cost scales with the training set size. It can be amortized across many query views, if the model is frozen.

In summary, VCURF achieves $O(Mr)$ per-view inference with $O(n)$ storage, while remaining training-free and black-box; alternatives either incur higher runtime (Entropy), higher storage (Ensemble), require white-box access (MCD/Entropy/FisherRF).

%% file: sup/qualitative.tex
\def\figwidth{0.23\textwidth}

\begin{figure*}[t]
\centering
\begin{tabular}{cccc}

\includegraphics[width=\figwidth]{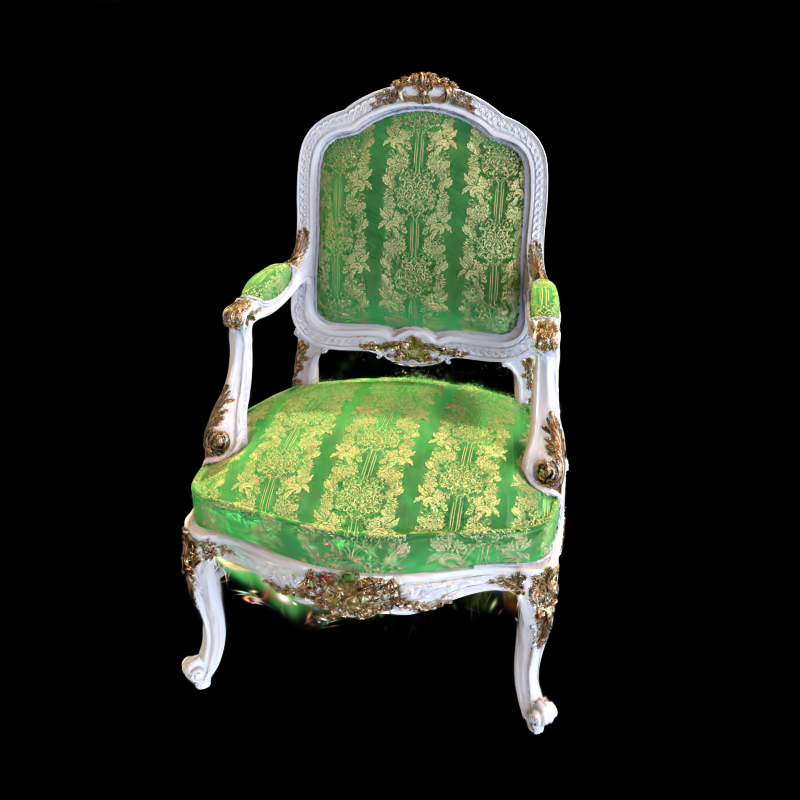} & 
\includegraphics[width=\figwidth]{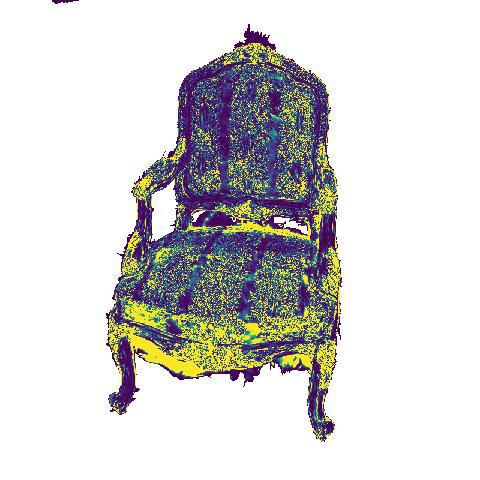} & 
\includegraphics[width=\figwidth]{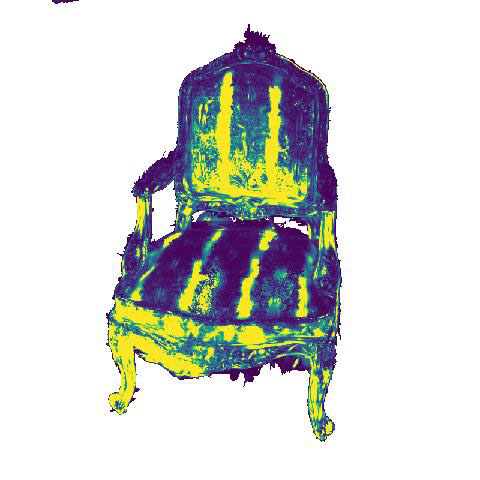} & 
\includegraphics[width=\figwidth]{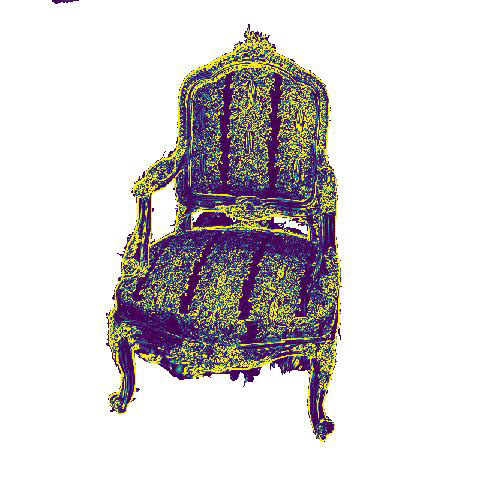} 
\\
Rendered image & RGB error & FisherRF &  VCURF \\

\includegraphics[width=\figwidth]{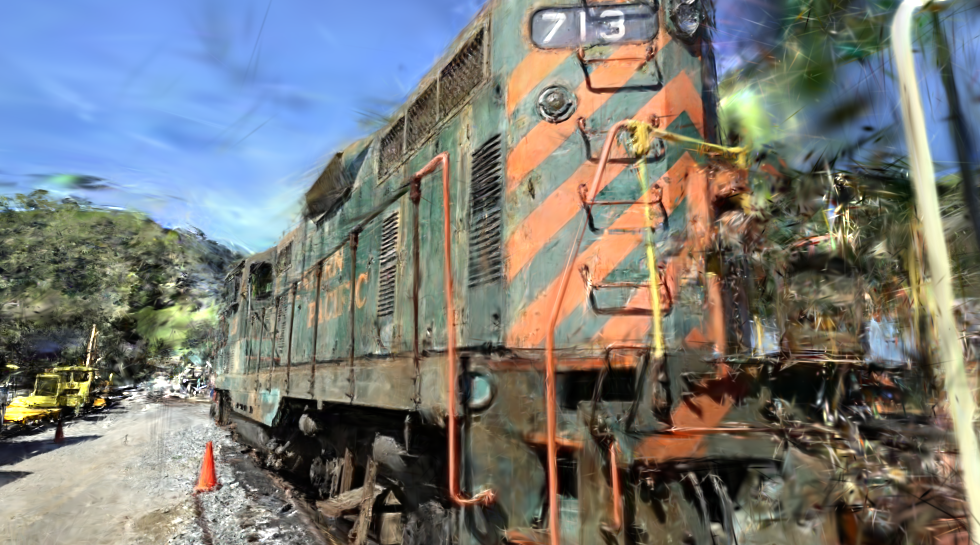} & 
\includegraphics[width=\figwidth]{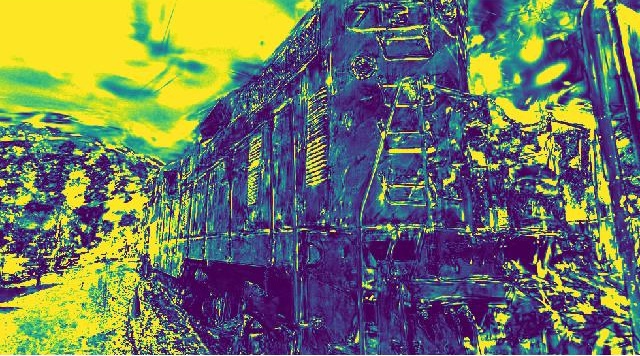} & 
\includegraphics[width=\figwidth]{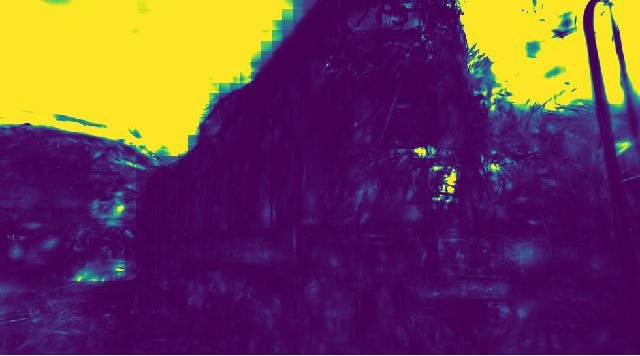} & 
\includegraphics[width=\figwidth]{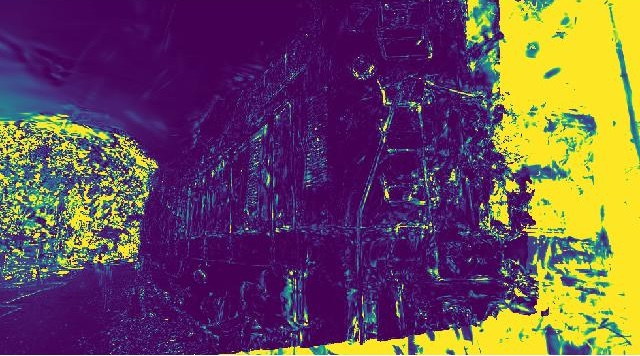} 
\\
Rendered image & RGB error & FisherRF &  VCURF \\

\includegraphics[width=\figwidth]{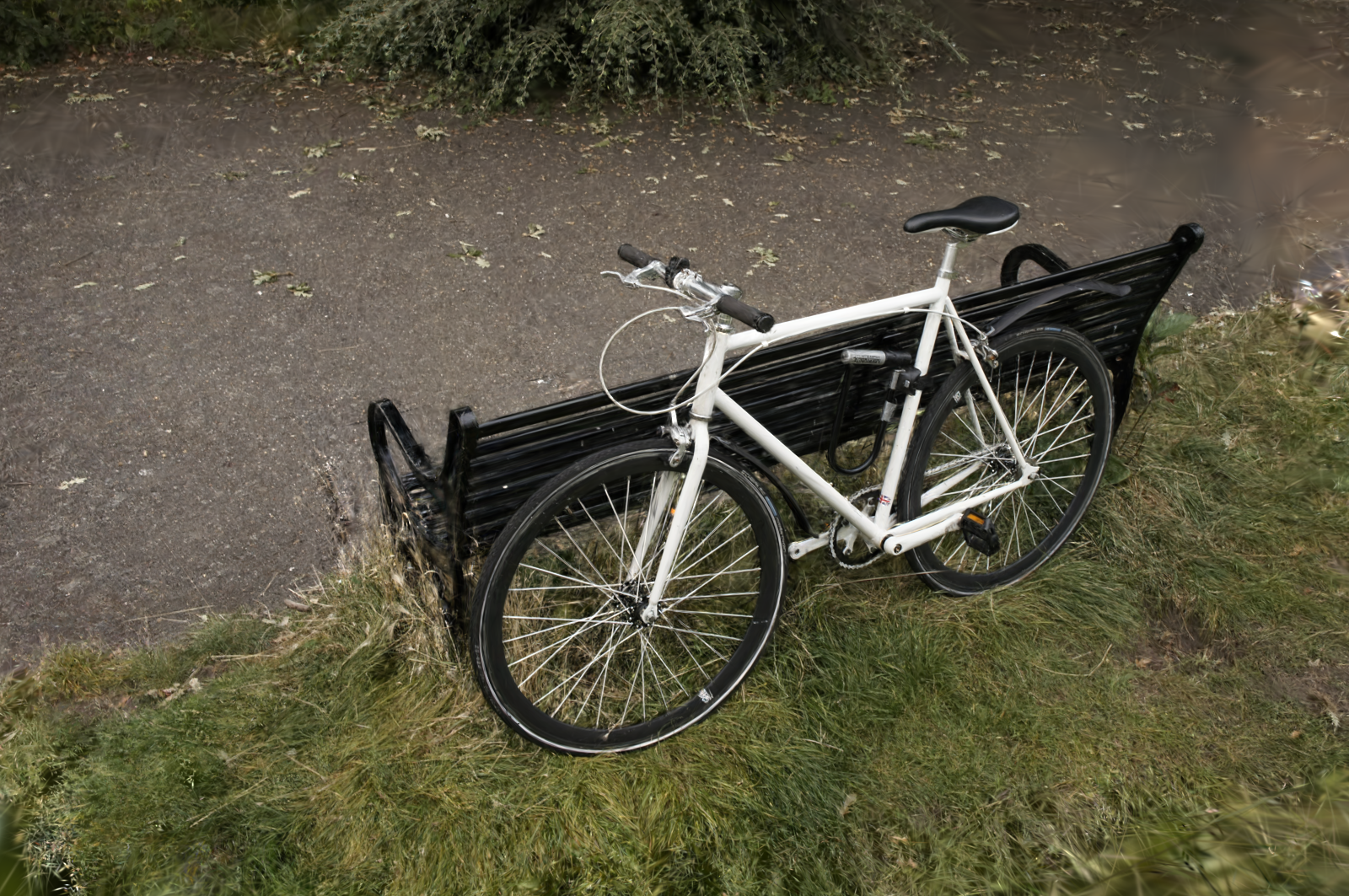} & 
\includegraphics[width=\figwidth]{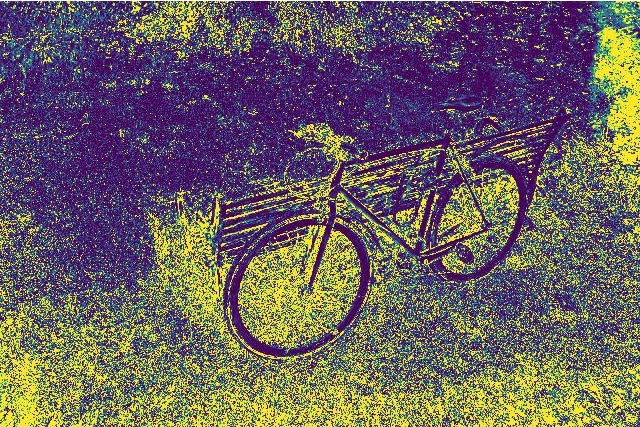} & 
\includegraphics[width=\figwidth]{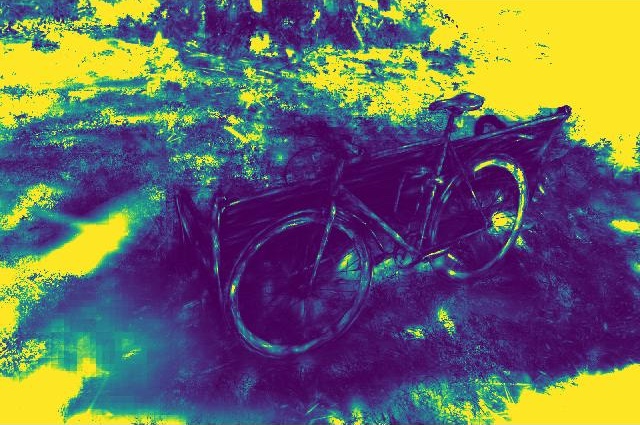} & 
\includegraphics[width=\figwidth]{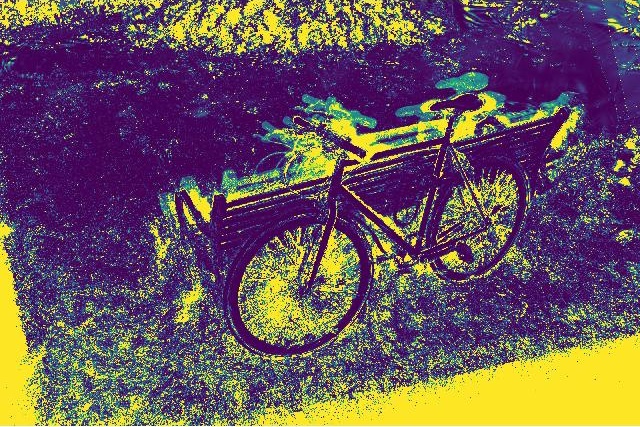} 
\\
Rendered image & RGB error & FisherRF &  VCURF \\

\includegraphics[width=\figwidth]{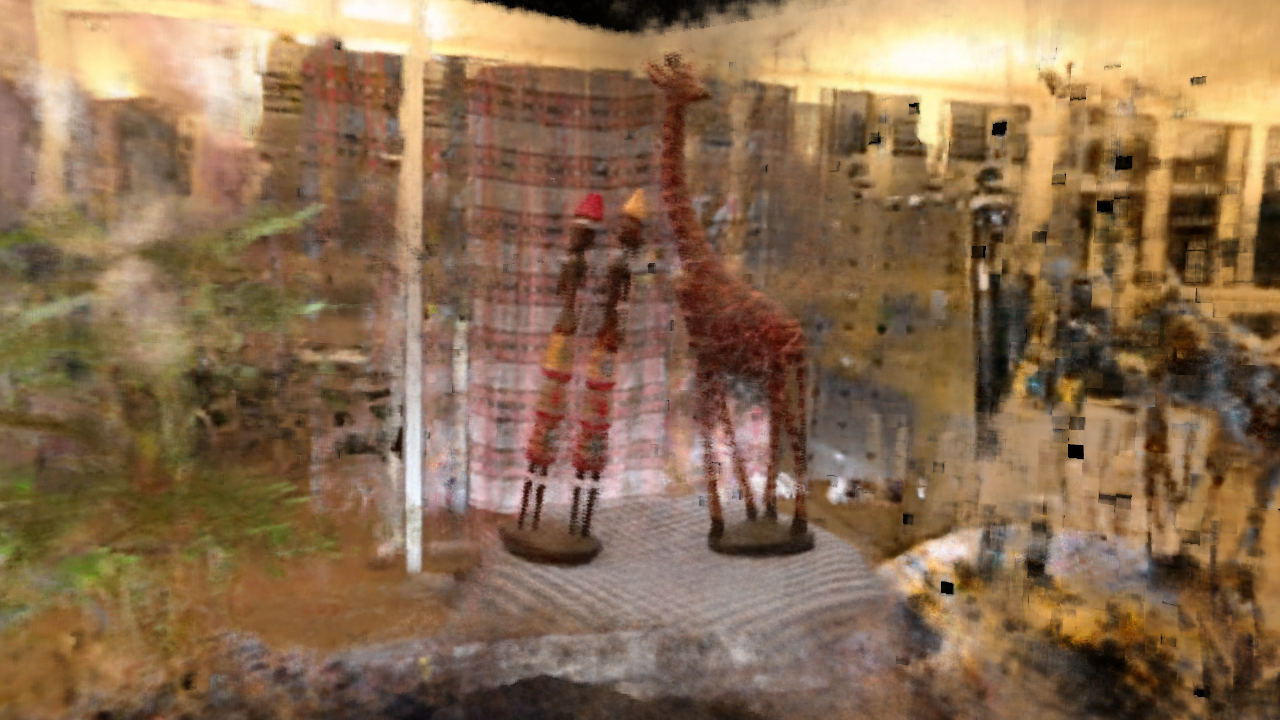} & 
\includegraphics[width=\figwidth]{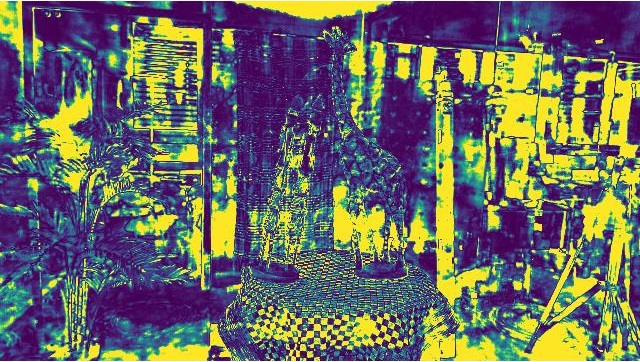} & 
\includegraphics[width=\figwidth]{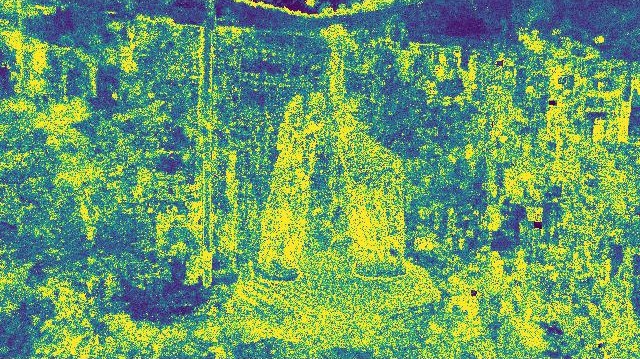} & 
\includegraphics[width=\figwidth]{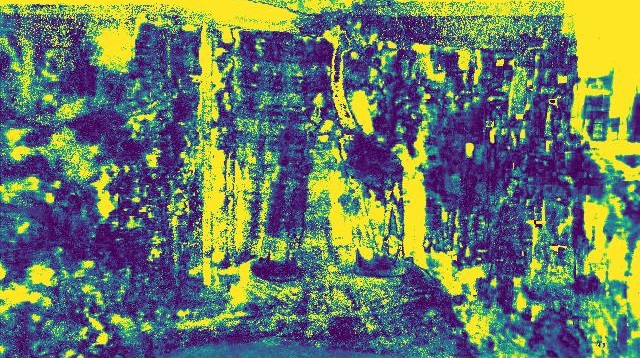} 
\\
Rendered image & RGB error & MCD-rgb &  VCURF \\

\includegraphics[width=\figwidth]{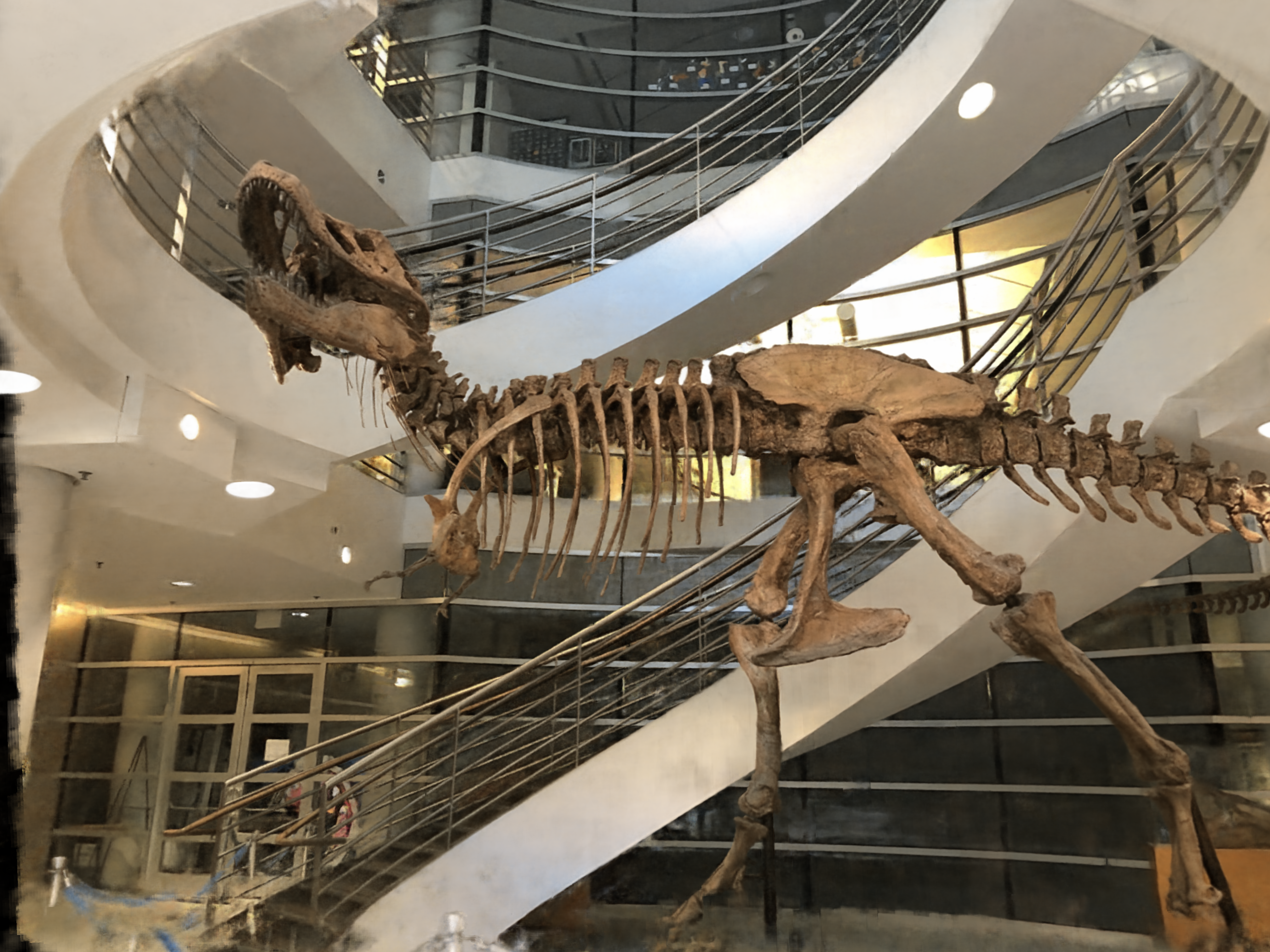} & 
\includegraphics[width=\figwidth]{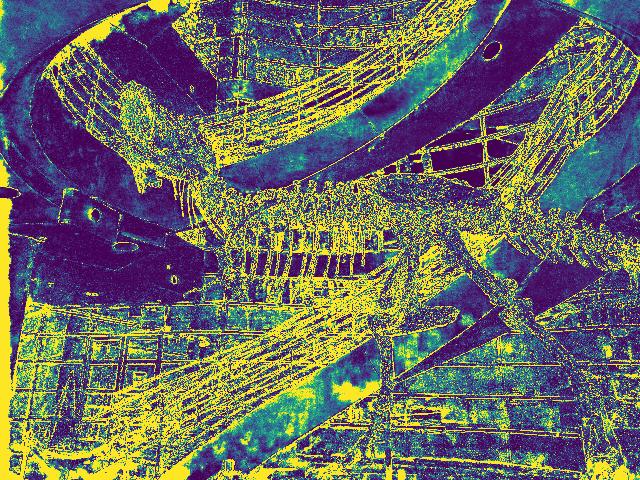} & 
\includegraphics[width=\figwidth]{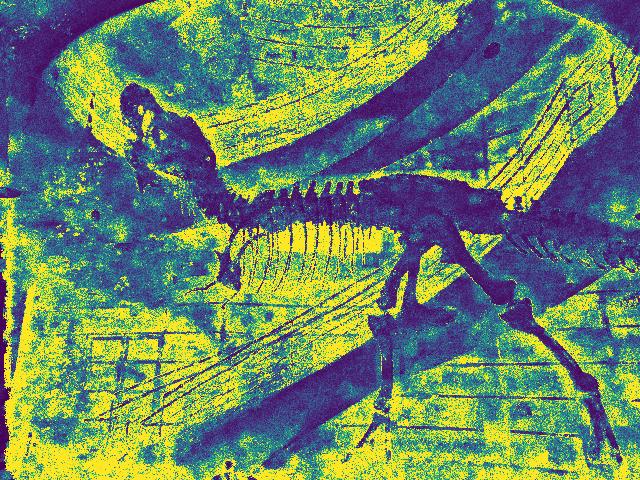} & 
\includegraphics[width=\figwidth]{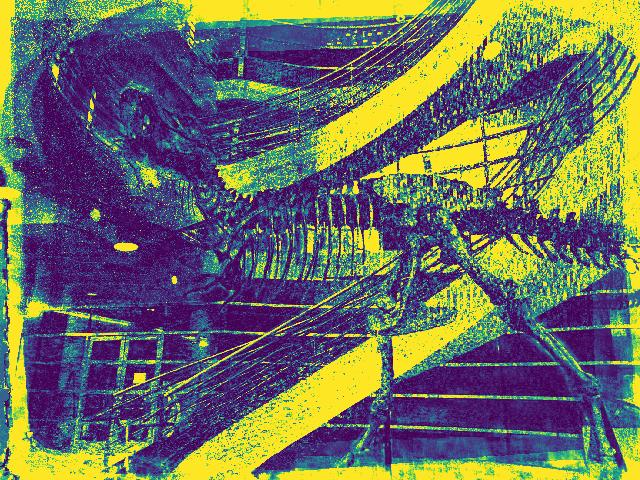} 
\\
Rendered image & RGB error & MCD-depth &  VCURF \\

\end{tabular}

\caption{Rendered images, error maps and corresponding uncertainty maps for one scene from each dataset. From top to bottom, they are: (1) \textit{chair} from Blender, (2) \textit{train} from Tank and Temples, (3) \textit{bicycle} from Mip-Nerf 360$^\circ$, (4) \textit{africa} from the Light Field, (5) \textit{t-rex} from Local Light Field Fusion. Following the colorbar of the main paper, yellow represents high values and blue represents low values.
The uncertainty maps of VCURF have better alignment with the error maps compared to the baselines. }
\label{fig:unc2}
\end{figure*}